\documentclass{article}

 \usepackage[eandd, final]{neurips_2026}
\setcitestyle{numbers,square,comma}

\usepackage[utf8]{inputenc} % allow utf-8 input
\usepackage[T1]{fontenc}    % use 8-bit T1 fonts
\usepackage{hyperref}       % hyperlinks
\usepackage{url}            % simple URL typesetting
\usepackage{amsfonts}       % blackboard math symbols
\usepackage{amsmath}
\usepackage{nicefrac}       % compact symbols for 1/2, etc.
\usepackage{microtype}      % microtypography
\usepackage{xcolor}         % colors
\usepackage{lipsum}
\usepackage{subcaption}
\usepackage{siunitx}
\usepackage{multirow}
\usepackage{caption}
\usepackage{wrapfig}
\usepackage{tabularx}
\usepackage{booktabs}
\usepackage[rightcaption]{sidecap}
\DeclareCaptionType{equ}[Equation][List of Equations]

\usepackage{booktabs} 

\usepackage{hyperref}

\usepackage{tikz}
\usetikzlibrary{arrows.meta, positioning, calc}

\usepackage[disable,linecolor=blue,backgroundcolor=blue!25,bordercolor=blue]{todonotes}

\usepackage{xspace}
\newcommand{\definesubstitution}[2]{\newcommand{#1}{#2\xspace}}

\definesubstitution{\dcc}{d_\text{CC}}
\definesubstitution{\normdcc}{\tilde{d}_\text{CC}}
\definesubstitution{\mindcc}{d_\text{CC}^{\min}}
\definesubstitution{\minnormdcc}{\tilde{d}_\text{CC}^{\min}}
\definesubstitution{\dcp}{d_\text{CP}}
\definesubstitution{\dwp}{d_\text{WP}}
\definesubstitution{\normdcp}{\tilde{d}_\text{CP}}
\definesubstitution{\mindcp}{d_\text{CP}^{\min}}
\definesubstitution{\minnormdcp}{\tilde{d}_\text{CP}^{\min}}
\definesubstitution{\nfp}{N_\text{FP}}
\definesubstitution{\ncoils}{N_\text{C}}
\definesubstitution{\curvature}{\kappa}
\definesubstitution{\normcurvature}{\tilde{\kappa}}
\definesubstitution{\maxcurvature}{\curvature^{\max}}
\definesubstitution{\maxnormcurvature}{\tilde{\curvature}^{\max}}
\definesubstitution{\normlength}{\tilde{l}}
\definesubstitution{\averagenormlength}{\tilde{l}_{mean}}
\definesubstitution{\fielderror}{\epsilon}
\definesubstitution{\logtenfielderror}{\log_{10}\left(\fielderror\right)}
\definesubstitution{\maxfielderror}{\fielderror^{\max}}
\definesubstitution{\regcoil}{\texttt{REGCOIL}~}
\definesubstitution{\desc}{\texttt{DESC}~}
\definesubstitution{\vmec}{\texttt{VMEC}~}
\definesubstitution{\mlp}{MLP}
\definesubstitution{\ensemblemlp}{EnsembleMLP}
\definesubstitution{\resmlp}{ResMLP}
\definesubstitution{\ensembleresmlp}{EnsembleResMLP}
\definesubstitution{\maskedmlp}{MaskedMLP}
\definesubstitution{\percoilmlp}{PerCoilMLP}
\definesubstitution{\ensemblemaskedmlp}{EnsembleMaskedMLP}
\definesubstitution{\ensemblepercoilmlp}{EnsemblePerCoilMLP}
\definesubstitution{\poincare}{Poincar\'e}

\newcommand{\project}{\texttt{CoilStellaration}}
\title{\project{}: A Dataset and Benchmark for Engineering-Aware Stellarator Coilset Generation}

\author{%
  Jan-Hendrik~Ewers\\ 
  Proxima Fusion\\
  Munich, Germany \\
  \texttt{jhewers@proximafusion.com} \\
  \And
  Santiago~A.~Cadena\\
  Proxima Fusion\\
  Munich, Germany \\
  \texttt{scadena@proximafusion.com} \\
  \And
  Nicolò~Foppiani\\
  Proxima Fusion\\
  Munich, Germany \\
  \texttt{nfoppiani@proximafusion.com} \\
  \And
  Andrea~Merlo\\
  Proxima Fusion\\
  Munich, Germany \\
  \texttt{amerlo@proximafusion.com} \\
}

\begin{document}

\maketitle

\listoftodos

\begin{abstract}

Stellarators are fusion devices that use externally generated,
three-dimensional magnetic fields to confine hot plasma and offer a promising route to commercial fusion.
Their electromagnetic coils pose a challenging design problem:
they must reproduce the target magnetic field while satisfying engineering constraints and remaining practical and cost-effective to manufacture.
Finding suitable geometries is a highly nonlinear inverse problem that generally has neither a closed-form nor a unique solution.
This problem lies at the heart of stellarator design.
We introduce CoilStellaration,
a dataset and benchmark for learning the mapping between coil geometries from target plasma magnetic fields and engineering requirements.
The dataset (\url{https://huggingface.co/datasets/proxima-fusion/coilstellaration}) contains 177,000 optimized filamentary coilsets targeting plasma equilibria from ConStellaration,
with sampled engineering requirements and achieved geometric and magnetic-field metrics. 
It spans multiple coil counts and field-period numbers,
supporting studies of both specialized and shared models across configurations. 
The benchmark asks models to predict a coilset in a single forward pass,
conditioned on a target equilibrium and requirements on coil–coil spacing, coil–plasma spacing, curvature, and magnetic-field error. 
We evaluate how often models satisfy all constraints and how much they violate those they fail to meet.
We show that magnetic-field accuracy remains the main challenge,
with learned models rarely satisfying all requirements simultaneously.
However,
using learned predictions to initialize conventional coil optimization can accelerate early progress.
We release baseline models and evaluation code to support reproducible research on both direct coil generation and learned optimization initializations. 
By lowering the barrier for machine learning and optimization researchers to engage with stellarator coil design, we aim to accelerate progress toward practical, cost-effective stellarator power plants.

\end{abstract}

\section{Introduction}
\label{sect:introduction}
Stellarators are magnetic confinement devices for hot plasma and currently stand as one of the primary contenders for commercial fusion energy on Earth~\cite{boozerStellaratorsFastPath2021,boozer2021carbondioxidefusionstellarators}. 
First proposed in 1951~\cite{spitzerStellaratorConcept1958},
they remained in the shadow of tokamaks for decades due to their engineering complexity. 
However,
recent breakthroughs have renewed their appeal as reactor candidates.
Discoveries of highly optimized stellarators~\cite{Landreman2022, Goodmanetal2023} demonstrate vastly improved plasma confinement. 
Wendelstein 7-X (W7-X) has demonstrated that optimized stellarators can be built and operated, and has confirmed several predicted confinement properties~\cite{Wolfetal2017, Dinklageetal2018, Wolfetal2019}.

The core element determining the quality of a stellarator's confinement field is its magnetic equilibrium.
This equilibrium is primarily generated by the set of coils,
the coilset,
which is a series of complex,
non-planar magnets that wind around the plasma.
The coilset dictates the majority of the stellarator's overall geometry and represents one of the most critical physical components that must be engineered.

The filamentary approximation represents each coil by a current-carrying curve.
It is useful for initial design, but neglects finite winding-pack dimensions, which can affect the magnetic field and require subsequent validation~\cite{singhOptimizationFinitebuildStellarator2020}.
This simplification allows designers to focus directly on key engineering constraints.
These constraints include the coil-to-plasma distance,
dictated by the need to fit internal components,
the minimum coil-to-coil distance,
driven by magnet technology and assembly space,
and the maximum coil curvature,
limited by the bending properties of the cable used in the coils.
Crucially,
these engineering limits are entirely dependent on the target machine.
The requirements for a physics experiment such as W7-X differ drastically from those of a commercial power plant.

Coilset design is an intrinsically complex problem.
Although computing the magnetic field from a given coilset geometry has a closed solution,
the problem of determining the coil geometry that produces a given field does not.
Therefore,
optimization is the primary method used in the field~\cite{strickler2002designing,zhuNewMethodDesign2017,giulianiDirectStellaratorCoil2022}.
Moreover,
according to physics first principles,
infinitely many coilset geometries can produce the same confining magnetic field.
They differ in geometrical features such as length,
curvature,
and relative spacing.

This complex inverse problem is well-suited for data-driven approaches.
A model can learn a conditional mapping from a target plasma boundary and engineering requirements to candidate coil geometries and currents.
This approach allows for rapid exploration of the design space and clearly outlines technological trade-offs.
It also enables fast iteration during the conceptual design phase,
when engineering requirements frequently shift.

In this paper we introduce \texttt{CoilStellaration}\footnote{Dataset: \url{https://huggingface.co/datasets/proxima-fusion/coilstellaration}}\footnote{Code: \url{https://github.com/proximafusion/coilstellaration}}, a comprehensive dataset of optimized stellarator coilsets.
Alongside the data,
we propose a machine learning benchmark for conditional coilset generation:
mapping a target plasma boundary and a specific set of engineering requirements to a viable coilset. 

This benchmark is part of a broader vision for data-driven stellarator design.
The \texttt{ConStellaration} dataset was a first step targeting stellarator equilibria (without coilset),
the inclusion of coilset (this work) represents a second foundational step,
and generative models for other critical reactor components will follow.
Ultimately,
this will enable joint optimization across the entire stellarator,
allowing engineers to balance trade-offs and requirements across entirely different physical subsystems.

\subsection{Background}
\label{sect:background}

\begin{figure}[b]
    \centering
    \includegraphics[width=\linewidth]{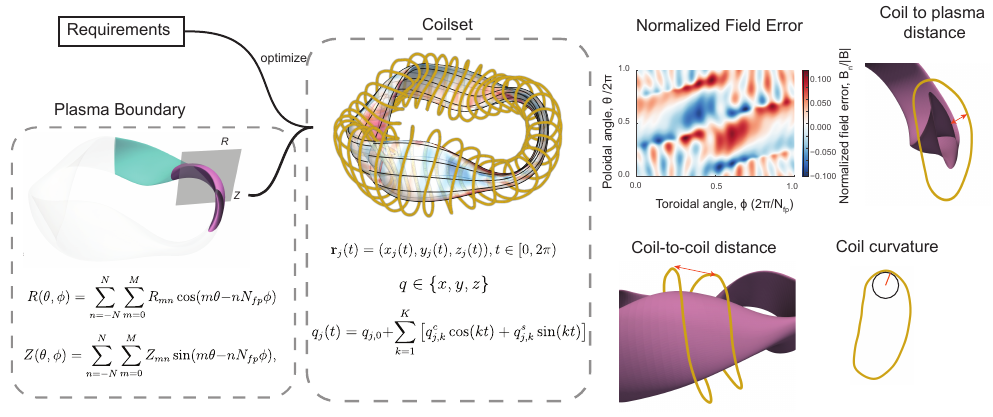}
    \caption{
    Left: The target plasma boundary,
    parameterized as a truncated Fourier surface in $\theta$ and $\phi$.
    Center: The generated coilset.
    Each filamentary coil is modeled as a closed 3D curve using a Fourier series in $t$,
    a parameter that runs along the curve.
    Solid black lines on the stellarator surface trace the equilibrium magnetic field lines,
    which are tangent to the boundary.
    Right:
    Key evaluation metrics.
    The heatmap displays the normalized magnetic field error across one plasma field period.
    The diagrams below illustrate the three physical engineering constraints used in this work:
    minimum coil-to-plasma distance,
    coil-to-coil spacing,
    and maximum coil curvature.
    }
    \label{fig:geometry-representation}
\end{figure}

Stellarator design traditionally follows a two-stage optimization process.
In the first stage,
the plasma equilibrium is optimized to satisfy specific physics properties.
This involves defining a 3D toroidal surface,
usually parameterized via a truncated Fourier series,
to act as the boundary condition for an ideal-magnetohydrodynamics (MHD) PDE model
(see \autoref{fig:geometry-representation}).
Physics codes like \texttt{VMEC++}~\cite{hirshman1983vmec,schilling2025numerics} solve these equations to determine the resulting magnetic field within that volume.
Through an iterative optimization loop,
the boundary shape is modified until the field reaches the target properties.
One important property is called omnigenity:
on average,
trapped particles do not drift outward or inward in the radial direction of the magnetic field.
As a result, they remain confined.
If a magnetic field is omnigenous and its strength contours form closed loops in the poloidal direction,
then that magnetic field is a quasi-isodynamic (QI) field.

The second stage,
and the primary focus of this work,
shifts to engineering.
The goal is to design the electromagnetic coils required to generate the magnetic field defined in the first stage.
This requires identifying coil geometries that accurately reproduce the target field while satisfying engineering constraints on curvature,
length,
and spacing.

% \begin{align}
% \mathbf{x}_{s}(\theta,\phi)
% =
% \bigl(
% R(\theta,\phi)\cos\phi,\,
% R(\theta,\phi)\sin\phi,\,
% Z(\theta,\phi)
% \bigr)
% \\
% R(\theta,\phi)
% =
% \sum_{n=-N}^{N}
% \sum_{m=0}^{M}
% R_{mn}
% \cos\!\left(m\theta-n\nfp\phi\right),
% \\
% Z(\theta,\phi)
% =
% \sum_{n=-N}^{N}
% \sum_{m=0}^{M}
% Z_{mn}
% \sin\!\left(m\theta-n\nfp\phi\right),
% \end{align}
% where $\nfp$ is the equilibrium number of field periods,
% $m$ and $n$ are the poloidal and toroidal mode numbers,
% respectively,
% and $R_{mn}, Z_{mn}$ are the set of Fourier coefficients describing the plasma boundary.
A \textit{flux surface} is a surface where magnetic field lines remain perfectly tangent.
The input plasma boundary defined in the first stage is usually a flux surface of the target equilibrium.
The primary optimization objective is to design coils that realize this surface by minimizing the quadratic flux:
the magnetic field component perpendicular (normal) to the plasma.
A successful design drives this normal component towards zero while simultaneously satisfying engineering constraints required for physical feasibility
(see \autoref{fig:geometry-representation}).

A Poincaré plot is a visual diagnostic used to verify the existence and quality of nested magnetic flux surfaces.
It is generated by tracing magnetic field lines over many toroidal transits:
each time a line intersects a fixed cross-sectional plane
(the Poincaré section),
a point is plotted.
In a successful design,
these intersections form smooth,
nested contours
(the nested flux surfaces).
Conversely,
poorly behaved field lines appear scattered,
forming chaotic regions or distinct "island chains" that might lead to degraded confinement (see \autoref{fig:metrics_correlation_with_coilset}).

\subsection{Related work}

Prior work has predominantly focused on datasets for stellarator plasma
equilibria. Large-scale databases of quasi-symmetric (QS) and QI
configurations have been constructed using near-axis expansion
frameworks~\cite{Landreman_2022,rodriguez2026near}.
Expanding beyond near-axis approximations, the
\texttt{ConStellaration}~\cite{cadenaConStellarationDatasetQIlike2025}
dataset provides diverse, QI-like configurations with ideal-MHD
equilibrium solutions computed for the entire plasma volume, in vacuum
and at finite $\beta$.

Regarding stellarator coilset datasets, the QUAsi-symmetric Stellarator
Repository (QUASR)~\cite{giulianiComprehensiveExplorationQuasisymmetric2025}
provides a database comprising nearly 370,000 quasi-axisymmetric (QA)
and quasi-helically (QH) symmetric devices paired with their optimized
coilsets. The dataset explores configurations with different aspect
ratios, rotational transforms, and coil lengths, subject to geometric
constraints on the coils.

Understanding how the target plasma boundary's geometry dictates the
complexity of the generating coilset is a central question for the
engineering feasibility of fusion power plants. Recent
work~\cite{kappelMagneticGradientScale2024,rodriguez2026estimating,pavone2026geodesic}
has shown how specific coil properties, such as coil-to-plasma distance
and coil non-planarity, can be captured by local magnetic field
properties.

CoilStellaration complements these resources by providing multiple
coilsets for prescribed QI-like plasma boundaries under different
coil-to-coil distance, coil-to-plasma distance, and curvature
requirements. This enables the exploration of trade-offs between
engineering requirements and magnetic field accuracy for a given
target boundary. A comparison of the three datasets is provided in
\autoref{app:dataset_comparison}.
Alongside the dataset, we introduce a standardized machine learning
benchmark and baseline architectures for conditional coilset generation.
This framework challenges models to directly map target plasma
boundaries and engineering requirements to viable coil geometries
and currents.

\section{Dataset}
\label{sect:dataset}
The dataset is generated through a comprehensive,
multi-stage pipeline designed to sample and optimize coilsets across a diverse range of engineering requirements.
The process begins by selecting a target plasma boundary and solving for its magnetic equilibrium.
We then sample a unique set of engineering constraints from a predefined distribution.
A preliminary coilset is initialized using \regcoil~\cite{landremanImprovedCurrentPotential2017} and subsequently refined through a constrained optimization using the \desc framework~\cite{dudtDESCStellaratorEquilibrium2020,paniciSurfaceCurrentOptimization2025}.
Finally,
the coilset is geometrically normalized,
its currents are rescaled to a target on-axis magnetic field strength,
and all relevant physical and engineering metrics are evaluated.

\paragraph{Equilibria filtering}
We retain QI-optimized target boundaries with a solved VMEC++ equilibrium and available QI diagnostics (\autoref{sect:appendix:boundary_filtering_details}).

% \subsubsection{Boundary filtering}
% \label{sect:dataset_generation:boundary_filtering}

% The \texttt{ConStellaration} dataset includes a wide diversity of configurations,
% including intermediate initializations that do not lead to fully resolved MHD equilibria.
% To ensure data quality,
% we implement a multi-stage filtering pipeline to isolate only those boundaries that map to fully QI-like,
% solved,
% force-balanced states.

% We first discard any configuration exhibiting non-convergence or runtime errors during the initial upstream optimization.
% Second,
% we require a minimum degree of QI to ensure the target configurations represent omnigenous equilibria.
% Finally,
% candidate boundaries are cross-referenced against high-fidelity solvers like VMEC++ to confirm they meet minimum force-balance criteria.
% A comprehensive breakdown of this filtering logic and the specific numerical thresholds used is provided in \autoref{sect:appendix:boundary_filtering_details}

\paragraph{Sampling pipeline}
\label{sect:dataset_generation:sampling_pipeline}
For each filtered boundary,
we sample a unique set of simplified engineering requirements.
The chosen values are informed from reference stellarator designs:
from small, compact devices to large, reactor-sized machines,
and they include both high-field designs using high-temperature superconductors (HTS) and low-field designs using low-temperature superconductors (LTS)
(see~\autoref{sect:appendix:ranges_engineering_requirements}).
These constraints are sampled from predefined ranges:
the number of unique coils per half-period ($\ncoils \in [4, 8]$),
the normalized minimum coil-to-plasma distance ($\normdcp = \dcp / a \in [0.5, 1.0]$),
the normalized minimum coil-to-coil distance ($\normdcc = \dcc / a \in [0.3, 0.7]$),
and the normalized maximum curvature ($\maxnormcurvature = a\maxcurvature \in [1.5, 5.0]$),
where $a$ denotes the minor radius of the target boundary.

% \begin{wrapfigure}{r}{0.6\linewidth}
% \vspace{-2em}
% \begin{minipage}{0.6\linewidth}

% \footnotesize
% \setlength{\abovedisplayskip}{3pt}
% \setlength{\belowdisplayskip}{3pt}
% \setlength{\abovedisplayshortskip}{2pt}
% \setlength{\belowdisplayshortskip}{2pt}
% \setlength{\jot}{2pt}

% \begin{equation*}
% \begin{aligned}
% \mathbf{x}^*
% = \arg\min_{\mathbf{x}}\quad
% & \mathcal{J}(\mathbf{x})
% = \frac12
% \left[
% \left\lVert \lambda\,\mathcal{J}_{L}(\mathbf{x}) \right\rVert_2^2
% +
% \left\lVert (1-\lambda)\,\mathcal{J}_{\Phi}(\mathbf{x}) \right\rVert_2^2
% \right]
% \\[-1pt]
% \text{s.t.}\quad
% & d^{cc}_i(\mathbf{x}) \ge d^{cc}_{\min}, \, i=1,\dots,N_c,
% \\
% & d^{cp}_i(\mathbf{x}) \ge d^{cp}_{\min}, \, i=1,\dots,N_c,
% \\
% & R^\kappa_i(\mathbf{x}) = 0, \, i=1,\dots,N_c,
% \\
% & \sum_{i=1}^{N_c} I_i = I^0 .
% \\
% & \mathcal{J}_{\Phi}(\mathbf{x})
% =
% \int_{\Sigma}
% \left[
% \mathbf{B}(\mathbf{r};\mathbf{x})
% \cdot
% \mathbf{n}(\mathbf{r})
% \right]^2
% \,\mathrm{d}S, \mathcal{J}_{L}(\mathbf{x})
% =
% \sum_{i=1}^{N_c} L_i^2(\mathbf{x}),
% \\
% & R^\kappa_i(\mathbf{x})
% =
% \frac{1}{L_i(\mathbf{x})}
% \int_{\Gamma_i}
% \left[
% \max\left\{
% |\kappa_i(\ell)|-\kappa_{\max},0
% \right\}
% \right]^2
% \,\mathrm{d}\ell .
% \end{aligned}
% \label{eq:coil-optimization-problem}
% \end{equation*}

% \end{minipage}
% \vspace{-1.2em}
% \end{wrapfigure}

\begin{wrapfigure}{r}{0.6\linewidth}
\vspace{-1em}
% The \fbox rule defines the border thickness
\setlength{\fboxrule}{0.8pt} 
\setlength{\fboxsep}{10pt} % Adjust padding between border and text
\fbox{
    \begin{minipage}{\dimexpr\linewidth-2\fboxsep-2\fboxrule\relax}
    
    \footnotesize
    \setlength{\abovedisplayskip}{3pt}
    \setlength{\belowdisplayskip}{3pt}
    \setlength{\abovedisplayshortskip}{2pt}
    \setlength{\belowdisplayshortskip}{2pt}
    \setlength{\jot}{2pt}
    
    \begin{equation*}
    \begin{aligned}
    \mathbf{x}^* =& \arg\min_{\mathbf{x}} \mathcal{J}(\mathbf{x}),\\
    \mathcal{J}(\mathbf{x}) =&
    \tfrac12\left[\lambda^2\mathcal{J}_L(\mathbf{x})
    +(1-\lambda)^2\mathcal{J}_{\Phi}(\mathbf{x})\right],\\
    \text{s.t.}\quad
    &d^{cc}_i(\mathbf{x})\ge d^{cc}_{\min},\quad
      d^{cp}_i(\mathbf{x})\ge d^{cp}_{\min},\\
    &R^\kappa_i(\mathbf{x})=0,\quad i=1,\dots,N_c,\qquad
      \sum_{i=1}^{N_c}I_i=I^0,\\
    &\mathcal{J}_L(\mathbf{x})=\sum_{i=1}^{N_c}
      \left(\frac{L_i(\mathbf{x})}{\ell_*}\right)^2,\\
    &\mathcal{J}_{\Phi}(\mathbf{x})=\sum_j W_j
      \left(\frac{\mathbf{B}_j(\mathbf{x})\cdot\mathbf{n}_j}
      {B_*R_*a_*}\right)^2,\\
    &R^\kappa_i(\mathbf{x})=
    \frac{\sum_j w_j\max\{|\kappa_{ij}|-\kappa_{\max},0\}^2}
         {\sum_j w_j}.
    \end{aligned}
    \end{equation*}
\refstepcounter{equation}
\small Equation~\theequation: Formulation of the coil optimization problem.
$L_i$ is coil length and $\mathbf{B}_j\cdot\mathbf{n}_j$ the normal field at surface node $j$.
$W_j$ and $w_j$ are surface and coil-parameter weights, respectively.
The fixed DESC normalization scales and weighting conventions are defined in \autoref{sect:appendix:objective_normalization}.
\label{eq:coil-optimization-problem}
\vspace{-0.8em}
\end{minipage}
}
\vspace{-1.2em}
\end{wrapfigure}

% \begin{equation}
% \begin{aligned}
% \mathbf{x}^* = \arg \min_{\mathbf{x}}\quad
% & \mathcal{J}(\mathbf{x}) \;=\; \tfrac12\bigl(\,\lVert\lambda\, \mathcal{J}_{L}(\mathbf{x})\rVert_2^{2}\;+\;\lVert(1-\lambda)\,\mathcal{J}_{\Phi}(\mathbf{x})\rVert_2^{2}\bigr) \\[4pt]
% \text{s.t.}\quad
% & d^{cc}_i(\mathbf{x})\;\ge\;d^{cc}_{\min}, && i=1,\dots,N_c,\\
% & d^{cp}_i(\mathbf{x})\;\ge\;d^{cp}_{\min}, && i=1,\dots,N_c,\\
% & R^\kappa_i = 0, && i=1,\dots,N_c,\\
% & \sum_{i=1}^{N_c} I_i \;=\; I^{0}. &&\\
% \end{aligned}
% \end{equation}

% where the quadratic flux error \cite{zhuDesigningStellaratorCoils2018} is defined as:
% \begin{equation}
% \mathcal{J}_{\Phi}(\mathbf{x}) \;=\; \int_{\Sigma}\bigl(\mathbf{B}(\mathbf{r};\mathbf{x})\!\cdot\!\mathbf{n}(\mathbf{r})\bigr)^{2}\,\mathrm{d}S(\mathbf{r}),
% \end{equation}

% the length penalty is defined as:
% \begin{equation}
%     \mathcal{J}_{L}(\mathbf{x}) \;=\; \sum_{i=1}^{N_c} L_i(\mathbf{x})^{2},
% \end{equation}

% and the curvature constraint \cite{krugerConstrainedStellaratorCoil2021} is defined as:
% \begin{equation}
% R^\kappa_i = \frac{1}{L_i(\mathbf{x})}\int_{\Gamma_i}\bigl(\max\!\bigl\{\lvert\kappa_i(\ell)\rvert-\kappa_{\max},\,0\bigr\}\bigr)^{2}\mathrm{d}\ell.
% \end{equation}

Coilsets are generated using a two-stage optimization approach. 
Each coil is represented by the Fourier coefficients describing its central filaments (see~\autoref{fig:geometry-representation}) and by the value of the current flowing through it.
We include Fourier modes up to $k=7$ with variable coil currents,
totaling 46 degrees of freedom per coil.
All optimizations fit a vacuum coil field to the prescribed target boundary. Matching the boundary does not by itself establish that the source equilibrium's pressure, current, or stability properties are reproduced.

In the first stage,
we use \regcoil to solve for a regularized surface current potential on a winding surface placed at a fixed distance $\dwp$ from the plasma.
To enhance the diversity of the resulting dataset,
we sample the plasma to winding surface distance,
$\dwp$,
independently from the coil-to-plasma distance $\dcp$.
Moreover,
the regularization strength is uniformly sampled between two mechanisms:
we either limit the maximum surface current density or the maximum magnetic field error.
A set of $N_c$ coils is then extracted from this optimized surface current potential to serve as the initial condition for the subsequent refinement.

In the second stage,
we refine the coilset directly using the \desc framework with the optimization equation defined in \autoref{eq:coil-optimization-problem}.
The optimization enforces the sampled minimum coil-to-plasma distance,
minimum coil-to-coil distance,
and maximum curvature as explicit constraints.
We minimize a weighted least-squares combination of normalized coil-length and normal-field residuals~\cite{zhuDesigningStellaratorCoils2018},
solved using an augmented Lagrangian method.

To further enhance the diversity of the dataset,
we optimize with different scalings of the design variables following an exponential scaling law with various norms~\cite{jang2026exponential}.
Following the optimization,
each coilset is normalized to a physical major radius of $1\,\mathrm{m}$. 
The coil currents are then rescaled to yield an average on-axis field strength of $B_{\mathrm{target}} = 1\,\mathrm{T}$.
% using the relation:
% \begin{equation}
%     I_{\mathrm{rescaled}}^i = I_{\mathrm{initial}}^i \frac{B_{\mathrm{target}}}{B_{initial}} = I_{\mathrm{initial}}^i \frac{1\,\mathrm{T}}{\langle \lVert \mathbf{B}(\mathbf{r}_{\mathrm{axis}}(\zeta)) \rVert \rangle_\zeta},
% \end{equation}
% where $I_{\mathrm{rescaled}}^i$ is the updated current for the $i$-th coil, and $I_{\mathrm{initial}}^i$ is the current of the $i$-th coil before rescaling.

\begin{figure}[t]
\centering
\begin{tikzpicture}[
  font=\small,
  >={Stealth[length=2mm]},
  node distance=5mm and 5mm,
  block/.style={
    draw, rounded corners, align=center,
    inner sep=4pt, minimum height=7mm, minimum width=15mm,
  },
  data/.style ={block, fill=gray!8},
  stage/.style={block, fill=blue!8, font=\small\bfseries},
  metric/.style={block, fill=orange!12, minimum height=7mm},
  output/.style  ={block, fill=green!12},
  arr/.style  ={->, thick},
]

% --- Top row: main dataflow ---------------------------------------------
\node[data]                       (eq)      {Equilibrium\\\scriptsize $\nfp\in [2,5] \}$, $\iota > 0$};
\node[stage, right=of eq]         (regcoil) {\regcoil\\warm start};
\node[stage, right=of regcoil]    (scale1)  {Rescale\\$B_0{=}1\,\mathrm{T}$};
\node[stage, right=of scale1]     (desc)    {Desc\\Aug.-Lagrangian\\BFGS};
\node[stage, right=of desc]       (scale2)  {Rescale\\$B_0{=}1\,\mathrm{T}$};
\node[output,   right=of scale2]     (ds)      {Dataset\\tuple};

% --- Bottom row: requirements + metrics ---------------------------------
\node[data,   below=7mm of eq]      (bnd) {Requirements\\bounds};
\node[data,   below=7mm of regcoil]          (req) {Sampled\\requirements};
\node[metric, below=3mm of scale1]  (m1)  {Metrics};
\node[metric, below=3mm of scale2]  (m2)  {Metrics};

% --- Main horizontal arrows ---------------------------------------------
\draw[arr] (eq)      -- (regcoil);
\draw[arr] (regcoil) -- (scale1);
\draw[arr] (scale1)  -- (desc);
\draw[arr] (desc)    -- (scale2);
\draw[arr] (scale2)  -- (ds);

% --- Requirements feed-in -----------------------------------------------
\draw[arr] (bnd) -- (req);
\draw[arr] (req.north) -- (regcoil.south);
\draw[arr] (req.east) -- ++(0,0) -| (desc.south);

% --- Metrics taps -------------------------------------------------------
\draw[arr] (scale1.south) -- (m1.north);
\draw[arr] (scale2.south) -- (m2.north);

% --- Metrics into dataset tuple -----------------------------------------
\draw[arr] (m2.east) -| (ds.south);
\draw[arr] (req.east) -| (ds.south);
\draw[arr] (eq.north) |- ++(1,0.4) -| (ds.north);

\end{tikzpicture}
\caption{End-to-end coilset generation pipeline. A sampled requirement vector
parameterizes a Regcoil warm start that is current-rescaled and then refined
by a constrained Desc optimization (length vs.\ quadratic-flux trade-off,
subject to coil--coil, coil--plasma, curvature, and total-current constraints).
Metrics on both the warm start and the refined coilset are emitted as part of the dataset tuple.}
\label{fig:pipeline}
\end{figure}
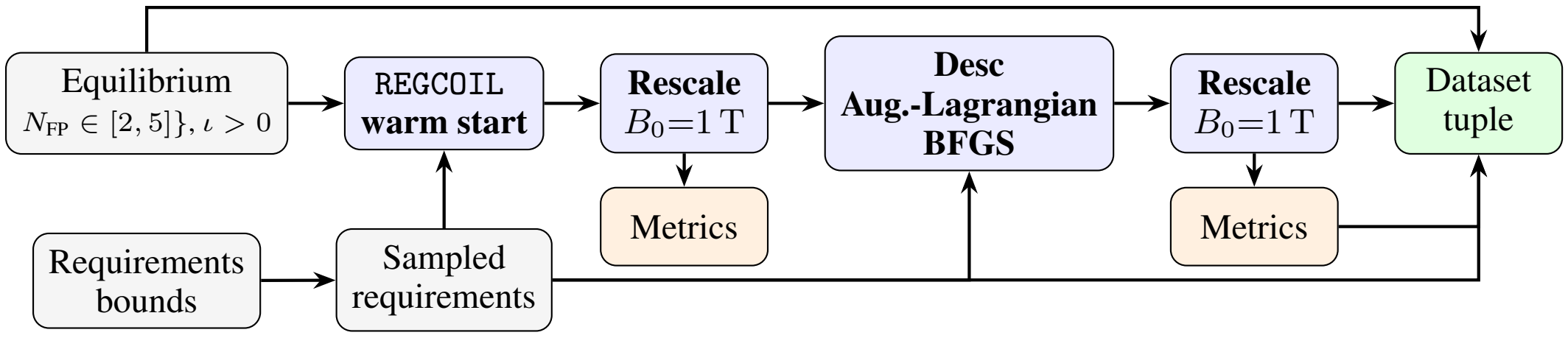

Eventually, for each generated coil pair, a complete metric record is computed 
(detailed in \autoref{sect:appendix:dataset_metrics}).
Each record includes the main physics and engineering diagnostics and reports the coilset's feasibility with respect to the initially sampled requirements.
A visual summary of the generation pipeline is illustrated in \autoref{fig:pipeline}.

Dataset generation is computationally intensive,
with the majority of the runtime consumed by \regcoil initialization and \desc refinement.
Computing the regularized surface current potential requires roughly $0.1$ CPUh,
while the more demanding filamentary optimization stage takes approximately $1.3$ CPUh.
The pipeline is executed on virtual machines configured with either \num{8} vCPU/\num{20} GB RAM or \num{16} vCPU/\num{40} GB RAM.

\paragraph{Dataset description}
\label{sect:dataset_description}

% \begin{wrapfigure}{r}{0.35\textwidth}
%   \vspace{-15pt}
%   \centering
%   \includegraphics[width=0.33\textwidth]{figures/dataset/nfp_vs_n_coils.pdf}
%   \caption{Distribution of the number of coils per half period ($\ncoils$) vs.\ number of field periods ($\nfp$). Colour scale is logarithmic.}
%   \label{fig:nfp_vs_n_coils}
%   \vspace{-10pt}
% \end{wrapfigure}

Each dataset sample contains one plasma boundary, one sampled requirement set, a \regcoil{} coilset, and, for 66.5\% of the samples, the refined \texttt{DESC} coilset (\autoref{sect:appendix:filter_breakdown} discusses the missing ones).
The dataset contains 266,549 samples, 177,341 \desc{} coilsets, and 14,496 unique plasma boundaries.
A more detailed breakdown is reported in \autoref{sect:appendix:detailed_dataset_breakdown}.
The dataset covers $\nfp\in\{2,3,4,5\}$ and $\ncoils\in\{4,\dots,8\}$, with $\nfp=3$ deliberately oversampled; not every combination survives quality filtering.

To evaluate the quality of these coilsets for further benchmark and model development, we focus on coilsets that reproduce the target boundary correctly.
The gold standard metric in the stellarator community is the average normalized field error, defined as:

\begin{equation}
\epsilon_{\mathrm{mean}} = \frac{\int_{\Sigma} |\mathbf{B}\cdot\mathbf{n}|/\|\mathbf{B}\|\,\mathrm{d}S}{\int_{\Sigma}\mathrm{d}S},
\end{equation}
where $\Sigma$ is the target plasma boundary surface, $\mathbf{B}$ is the magnetic field generated by the coilset, $\mathbf{n}$ is the unit normal vector to the surface, and $\mathrm{d}S$ is the differential area element.
Values below 1\% are typically considered good enough to reproduce the target field.
Because the \regcoil coilset almost never achieve this value, we restrict every further analysis to \desc coilset only.

\begin{figure}[htbp]
  \centering
  \includegraphics[width=\textwidth]{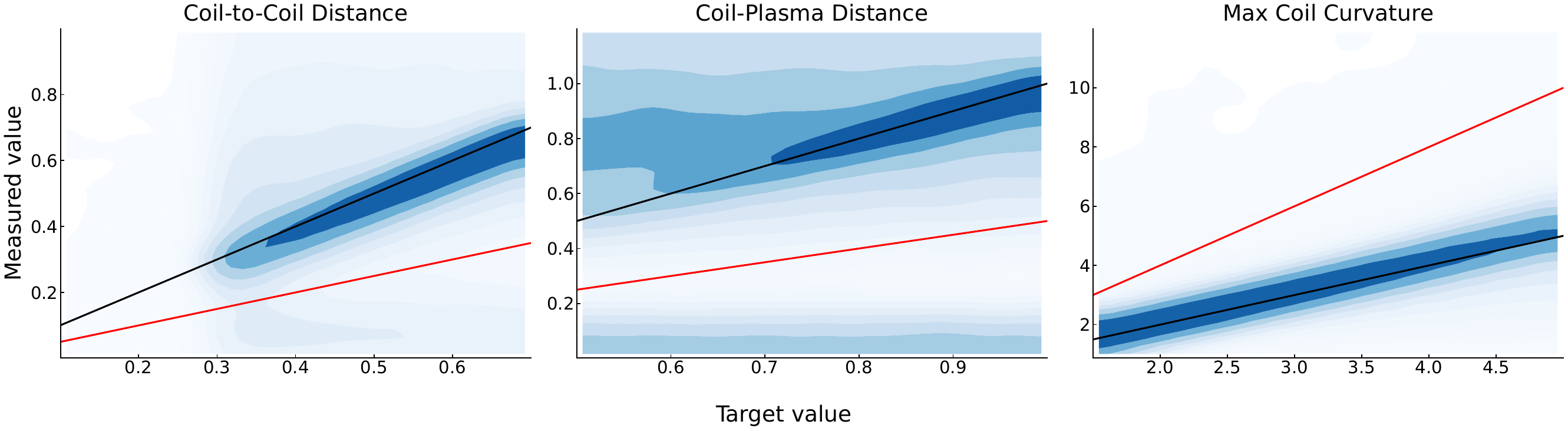}
  \caption{Measured vs.\ target values for the three geometric constraints.
           Colour scale is logarithmic.
           Black line: requirement equals measured.
           Red line: selection cut threshold.}
  \label{fig:target_vs_measured}
\end{figure}

\begin{SCfigure}[][htbp]
  \centering
  \includegraphics[width=0.7\textwidth]{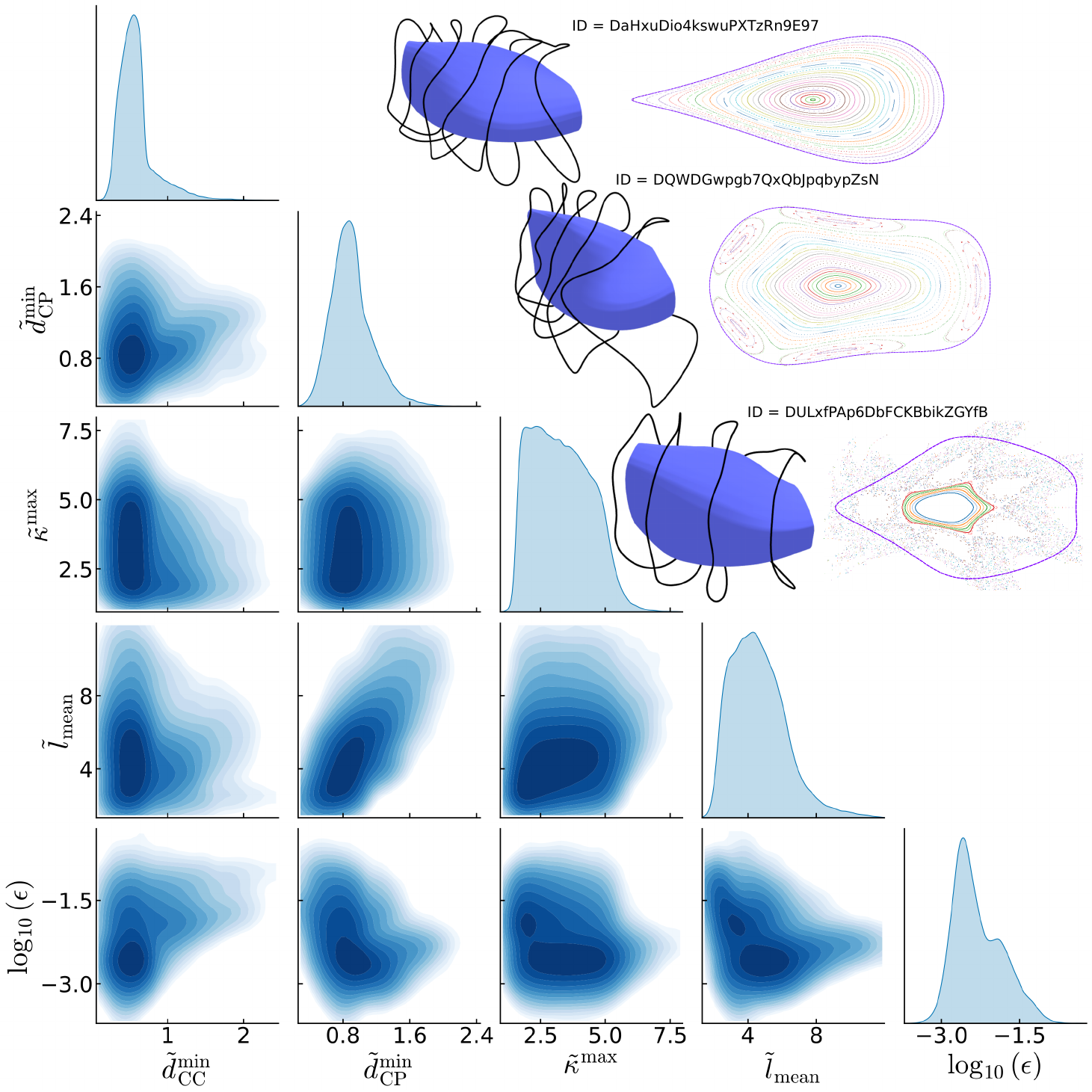}
  \caption{
    The filtered dataset spans a broad range of geometric metrics and magnetic-field errors. The pairwise plots show how these quantities vary together.
    The dataset covers a large range in $\minnormdcc$, $\minnormdcp$, $\maxnormcurvature$, $\averagenormlength$, and field error.
    The three coilsets are sampled at increasing field error $\epsilon$, reproducing the target plasma boundary with different accuracy.
    The Poincare plot of the first coilset presents a clean set of flux surface till the plasma boundary, the second one has flux surfaces in the core, but degraded at the boundary, while the third one displays almost no flux surface.
  }
  \label{fig:metrics_correlation_with_coilset}
\end{SCfigure}

We also evaluated how well the optimization pipeline met the requirements.
\autoref{fig:target_vs_measured} shows measured quantities against the requirements, outlining two distinct populations:
the samples that sit around the measured=target line (black) and the sample for which the measure value is close to zero for $\minnormdcc$ and $\minnormdcp$ or very large for $\maxnormcurvature$.
This second population contains coilsets for which the optimization completely missed the requirements, which we filter out by requiring $e_{\minnormdcc} > -0.5$ and 
$e_{\minnormdcp} > -0.5$ and 
$e_{\maxnormcurvature} < 1.0$,
with the relative error $e$ defined as $e = (x_{\text{measured}} - x_{\text{target}}) / x_{\text{target}}$.
The selection filters out 21.8\% of the \desc coilsets, with 7.6\% for which the metrics calculation failed, leaving a clean dataset of coilsets spanning a wide range of metrics \autoref{fig:metrics_correlation_with_coilset}.
A more detailed breakdown of the filter efficiency is reported in \autoref{sect:appendix:filter_breakdown}.

\section{Benchmark \& Baselines}
\label{sect:benchmark_baselines}

We introduce a benchmark to test if a model can generate engineerable coilsets in a single forward pass.
The difficulty of this task is exactly what makes it valuable:
traditional optimization is computationally expensive.
A sufficiently accurate model could reduce the cost of exploring candidate coil designs by predicting a coilset in a single forward pass.

\subsection{Benchmark}
\label{sect:benchmark}

\paragraph{Task.}
Given a target plasma boundary $\mathbf{b}$ and an engineering requirements
vector $\mathbf{r} = (\minnormdcc, \minnormdcp, \maxnormcurvature, \maxfielderror)$,
a model $f_\theta$ must produce a coilset $\mathbf{c} = f_\theta(\mathbf{b}, \mathbf{r})$
in a single forward pass --- no time-consuming optimization is allowed at inference. The first two requirements are lower bounds the achieved value must
exceed; the latter two are upper bounds it must not exceed. We re-evaluate the
achieved metrics $g_i$ on $\mathbf{c}$ via the same \desc{} filamentary pipeline
used during dataset generation (\autoref{sect:dataset}) and grade
them against $\mathbf{r}$.  

\paragraph{Per-metric soft-feasibility.}
For each metric we compute a normalized violation $v_i \ge 0$ that is zero iff
the constraint holds:
\begin{equation}
\label{eq:violation}
\begin{aligned}
v_i &= \max\!\left(0,\frac{g_i-\tau_i}{\max(|\tau_i|,10^{-6})}\right)
&&\text{(upper bound)},\\
v_i &= \max\!\left(0,\frac{\tau_i-g_i}{\max(|\tau_i|,10^{-6})}\right)
&&\text{(lower bound)}.
\end{aligned}
\end{equation}
A saturating exponential maps each violation to a soft-feasibility score on
$[0, 1]$,
\begin{equation}
\label{eq:soft-score}
s_i = \exp\!\bigl(-v_i / \alpha_i\bigr) \in [0, 1],
\end{equation}
where $\alpha_i$ is a per-metric tolerance frozen before any baseline run
(\autoref{tab:appendix:tolerances}). Saturation prevents trading surplus on one
metric against slack on another. The score decreases continuously as a violation grows, and the common $[0,1]$ scale makes the four metrics comparable. When the denominator floor is inactive, the tolerance has a physical reading: violating constraint $i$ by
$\alpha_i {\cdot} 100\%$ of the threshold reduces $s_i$ to $1/e \approx 0.37$.

\paragraph{Aggregation across metrics.}
We collapse the four per-metric scores into a per-instance score with an
unweighted geometric mean,
\begin{equation}
\label{eq:geomean}
S = \Bigl(\prod_{i=1}^{4} s_i\Bigr)^{1/4}.
\end{equation}
Geometric mean is unforgiving by design: any single $s_i \approx 0$ collapses
$S$, irrespective of the other three. This matches engineering reality:
a coilset that fails one requirement is not feasible.
We avoid the arithmetic
mean as the headline because an arithmetic $\bar S = 0.8$ can hide ``feasible
on three, collapsed on one'', which is the failure mode the benchmark must
surface.

\paragraph{Aggregation across test instances.}
For a test set $\mathcal{T} = \{(\mathbf{b}_n, \mathbf{r}_n)\}_{n=1}^N$ with
per-instance scores $\{S_n\}$, we report the mean $\bar S$ as the headline
ranking column, the median as a typical-instance diagnostic, the
soft-feasibility rate $\Pr[S_n \ge 0.9]$, and the strict-feasibility rate
$\Pr[\forall i\!:\! v_{n,i} \le \num{1e-3}]$. Together these distinguish models that
are consistently feasible from those that nail a few easy instances and
collapse on the rest. Strict feasibility allows normalized violation up to $10^{-3}$; this is $0.1\%$ relative tolerance when the denominator floor is inactive.

\paragraph{Split by boundary.} A training and evaluation data split is constructed to be disjoint by the plasma boundary: each boundary has a unique ConStellaration-based hash (the ID) on which a $95\%/5\%$ split was done. The hash-binning guarantees that there is no boundary overlap between the datasets, even though there are multiple entries per boundary. This resulted in 253,000 and 13,000 rows in the training and evaluation splits respectively. 
\paragraph{Two tracks.}
A test instance has both a continuous component ($\mathbf{b}, \mathbf{r}$) and
a discrete topology $(\nfp, \ncoils)$. We expose two tracks. The
\textit{fixed-shape} track evaluates models on a single discrete topology
$(\nfp{=}3,\, \ncoils{=}5)$ --- the most common configuration in the dataset,
oversampled at generation time --- to test whether models can specialize on
one of the recurring discrete choices in real reactor design. The
\textit{variable-shape} track evaluates models across the full range of
$(\nfp, \ncoils)$ in the training distribution, testing whether one model can
handle the joint topology--continuous space. 
\paragraph{Stratification by requirement tightness.}
For each boundary in the evaluation split,
we extracted multiple requirement vectors
derived from the metrics our data-generation pipeline achieved on different
\desc{} reference coilsets. Because the pipeline is biased toward low
normalized field error, we stratify each track into three tiers ---
\textbf{tight}, \textbf{medium}, \textbf{loose} --- by tertiles of the mean
normalized field error $\langle\varepsilon_B\rangle$ that \desc{} achieved on
the reference coilset: with $q_{1/3}$ and $q_{2/3}$ the corresponding quantiles
over all generated instances, an instance is tight if
$\langle\varepsilon_B\rangle \le q_{1/3}$, medium if
$q_{1/3} < \langle\varepsilon_B\rangle \le q_{2/3}$, and loose otherwise.
Tiers are defined by the reference coilset's mean normalized field error.
The field-error requirement instead bounds the maximum sampled normalized field error.
The tiers do not jointly order all four requirements.

\paragraph{Benchmark challenges.}
The benchmark exposes the joint difficulty of (i)~conditioning a generator on
a continuous engineering vector $\mathbf{r}$, (ii)~discrete topology choices
$(\nfp, \ncoils)$, and (iii)~physical constraints whose feasibility surfaces
in 3D coil geometry are strongly non-convex. The geometric-mean aggregation
is intolerant of partial feasibility, which is the right primitive for
downstream engineering use. Stratification by $\langle\varepsilon_B\rangle$ lets us read
off how feasibility scales with constraint tightness rather than collapse to a
single average. The single-forward-pass constraint cleanly separates the
benchmark from classical optimization, so progress is unambiguously progress
on amortized reactor-coil design.

\subsection{Baselines}
\label{sect:baselines}

We propose four baselines, all sharing a bidirectional
spectral-scaling interface that maps Fourier-parameterized boundaries and
coilsets to flat vectors and back (full architecture details in
\autoref{sect:appendix:model_config_and_architecture}):
\textbf{\mlp}, a feed-forward net with GELU activations producing both spatial
coil arrays and scalar currents;
\textbf{\resmlp}, pre-norm residual blocks with split heads (a primary linear
head for geometry and a zero-initialized currents head conditioned on the
predicted geometry);
and \textbf{\ensemblemlp{}} / \textbf{\ensembleresmlp{}}.
Each ensemble contains five independently initialized networks trained on shared minibatches.
We average their individual supervised losses during training and their predicted coil coefficients and currents at inference.
We repeat each single-model or ensemble run with three seeds and report variation across those runs, on $2{,}771$ fixed-shape instances. Each candidate took approximately $0.5$\,h to train on a machine with 32 CPU cores and 64\,GB of RAM.

\begin{wrapfigure}{r}{0.6\linewidth}
  \centering
  \includegraphics[width=\linewidth]{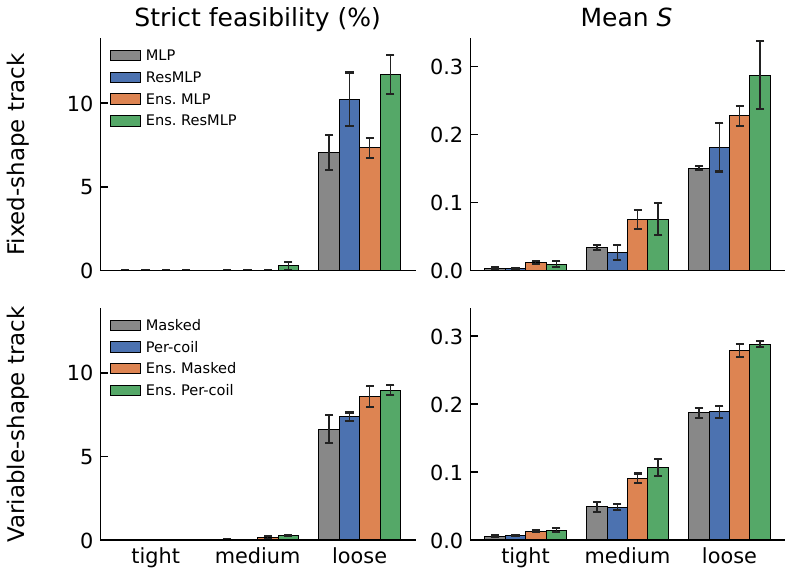}
  \caption{Strict feasibility (\%) and mean score $S$ by requirement tier
  (tight / medium / loose), for the fixed-shape track (top) and the
  variable-shape track (bottom). Each track shows its own four architectures.
  Bars are the mean over three training seeds with $\pm 1\sigma$ error bars.
  Across both tracks and all architectures, strict feasibility is $0\%$ on the
  tight tier and near-zero on the medium tier: nearly all strictly feasible
  coilsets fall in the loose tier. Strict feasibility uses the benchmark
  tolerance $v_i \le \num{1e-3}$.}
  \label{fig:baselines_by_tier}
\end{wrapfigure}

\paragraph{All baselines are far from saturated.}
The best mean score $\bar S$ is only $0.13$, and even the best strict-feasibility
rate is only $4.4\%$ (\autoref{tab:baselines}). These rates come almost entirely
from the loosest requirements: on the tight tier not a single baseline produces one strictly feasible coilset
(\autoref{fig:baselines_by_tier}). The low feasibility rates leave substantial room for improvement.

\paragraph{Difficulty grows sharply with requirement tightness.}
\autoref{fig:baselines_by_tier} stratifies strict feasibility and mean $S$ by
the tight/medium/loose tiers. Strict feasibility is $0\%$ on the tight tier for
every architecture on both tracks and essentially zero on the medium tier
($<0.3\%$); nearly all strictly feasible coilsets fall in the loose tier
(fixed-shape: $7\text{--}12\%$; variable-shape: $7\text{--}9\%$). Mean $S$
follows the same ordering, dropping by more than an
order of magnitude from the loose to the tight tier. The aggregate leaderboard
numbers therefore reflect loose-tier performance. No evaluated baseline achieves strict feasibility in the tight reference-field-error tier.

\paragraph{\ensembleresmlp{} is the strongest baseline.}
Two effects compound (\autoref{tab:baselines}). Ensembling raises the mean score
for both backbones, and the residual backbone adds feasibility: both residual
variants (\resmlp{}, \ensembleresmlp{}) beat their plain-\mlp{} counterparts on
strict feasibility. Combining the two, \ensembleresmlp{} attains the best mean
$\bar S$, soft- and strict-feasibility rates; \ensemblemlp{} retains the best
median $S$. The difference in mean score between MLP and ResMLP is small relative to the variation across the three training runs.
Field-error-only changes preserve the ranking; uniform scaling can change it
(\autoref{sect:appendix:alpha_sensitivity}).

\begin{table*}[t]
\centering
% \tiny

\begin{minipage}[t]{0.49\textwidth}
\vspace{0pt}
\centering
\caption{Leaderboard on the fixed-shape track ($\nfp{=}3$, $\ncoils{=}5$, $N{=}2{,}771$), mean $\pm\,1\sigma$ over three training seeds, ranked by $\bar S$. ``Soft'' is $\Pr[S\!\ge\!0.9]$ ($\%$); ``Strict'' is $\Pr[\forall i\!:\! v_i{\le}\num{1e-3}]$ ($\%$). Best per column in \textbf{bold}.}
\label{tab:baselines}

\resizebox{\linewidth}{!}{%
\begin{tabular}{@{}lcccc@{}}
\toprule
Architecture & $\bar S$ & $\mathrm{med}\,S$ & Soft (\%) & Strict (\%) \\
\midrule
\ensembleresmlp & $\mathbf{0.133\pm0.027}$ & $5.9\!\times\!10^{-3}$          & $\mathbf{6.5\pm0.7}$ & $\mathbf{4.4\pm0.5}$ \\
\ensemblemlp    & $0.111\pm0.003$          & $\mathbf{7.1\!\times\!10^{-3}}$ & $3.5\pm0.2$          & $2.7\pm0.2$          \\
\resmlp         & $0.076\pm0.017$          & $2.1\!\times\!10^{-4}$          & $5.2\pm0.9$          & $3.8\pm0.6$          \\
\mlp            & $0.067\pm0.002$          & $4.9\!\times\!10^{-4}$          & $3.3\pm0.5$          & $2.6\pm0.4$          \\
\bottomrule
\end{tabular}%
}
\end{minipage}
\hfill
% ---------- Variable-shape ----------
\begin{minipage}[t]{0.49\textwidth}
\vspace{0pt}
\centering
\caption{Leaderboard on the variable-shape track ($N{=}8{,}058$ instances spanning $\nfp\in\{2,\dots,5\}$, $\ncoils\in\{4,\dots,8\}$), mean $\pm\,1\sigma$ over three training seeds, ranked by $\bar S$. Columns as in \autoref{tab:baselines}. Best per column in \textbf{bold}.}
\label{tab:baselines_variable}

\resizebox{\linewidth}{!}{%
\begin{tabular}{@{}lcccc@{}}
\toprule
Architecture & $\bar S$ & $\mathrm{med}\,S$ & Soft (\%) & Strict (\%) \\
\midrule
\ensemblepercoilmlp & $\mathbf{0.137\pm0.006}$ & $\mathbf{1.8\!\times\!10^{-2}}$ & $\mathbf{4.6\pm0.2}$ & $\mathbf{3.1\pm0.1}$ \\
\ensemblemaskedmlp  & $0.128\pm0.005$          & $1.4\!\times\!10^{-2}$          & $4.3\pm0.3$          & $2.9\pm0.2$          \\
\percoilmlp         & $0.082\pm0.005$          & $2.3\!\times\!10^{-3}$          & $3.6\pm0.1$          & $2.5\pm0.1$          \\
\maskedmlp          & $0.081\pm0.005$          & $2.3\!\times\!10^{-3}$          & $3.3\pm0.4$          & $2.2\pm0.3$          \\
\bottomrule
\end{tabular}%
}
\end{minipage}

\end{table*}

% \begin{table}[t]
% \tiny
% \centering
% \caption{Leaderboard on the fixed-shape track ($\nfp{=}3$, $\ncoils{=}5$, $N{=}2{,}771$), mean $\pm\,1\sigma$ over three training seeds, ranked by $\bar S$. ``Soft'' is $\Pr[S\!\ge\!0.9]$ ($\%$); ``Strict'' is $\Pr[\forall i\!:\! v_i{\le}\num{1e-3}]$ ($\%$). Best per column in \textbf{bold}.}
% \label{tab:baselines}
% \begin{tabular}{@{}lcccc@{}}
% \toprule
% Architecture & $\bar S$ & $\mathrm{med}\,S$ & Soft (\%) & Strict (\%) \\
% \midrule
% \ensembleresmlp & $\mathbf{0.133\pm0.027}$ & $5.9\!\times\!10^{-3}$          & $\mathbf{6.5\pm0.7}$ & $\mathbf{4.4\pm0.5}$ \\
% \ensemblemlp    & $0.111\pm0.003$          & $\mathbf{7.1\!\times\!10^{-3}}$ & $3.5\pm0.2$          & $2.7\pm0.2$          \\
% \resmlp         & $0.076\pm0.017$          & $2.1\!\times\!10^{-4}$          & $5.2\pm0.9$          & $3.8\pm0.6$          \\
% \mlp            & $0.067\pm0.002$          & $4.9\!\times\!10^{-4}$          & $3.3\pm0.5$          & $2.6\pm0.4$          \\
% \bottomrule
% \end{tabular}
% \end{table}

\paragraph{Field error is the dominant failure mode.}
Decomposing $S$ into its four soft-score components
(\autoref{tab:appendix:per-metric}) shows the geometric-mean collapse is
overwhelmingly driven by the field-error term: median normalized field-error violations are $2.1$--$2.9$ across all architectures,
driving $s_\epsilon$ into the $10^{-5}\!-\!10^{-4}$ range. Both ensembles have median soft scores near one for coil--plasma distance and curvature, while every architecture has a positive median coil--coil spacing violation.
Field-error reproduction is the highest-leverage research direction we identify
for this benchmark.

% \paragraph{Qualitative behaviour under requirement perturbation.}
\todo[inline]{Narrow full-width figure: predicted coilsets from one model on a fixed boundary as we sweep one requirement (e.g. $\minnormdcc$) across its range, so the deformation is visually apparent.}
% \begin{figure*}[t]
%   \centering
%   \includegraphics[width=\linewidth]{figures/coilset_requirement_sweep.pdf}
%   \caption{Predicted coilsets as a function of a single swept requirement,
%            holding the boundary fixed.}
%   \label{fig:coilset_requirement_sweep}
% \end{figure*}

% Comment from AM.
% Hypothesis for why the model does not achieve good results:
% - not enough data -> plot feasibility over number of samples in the training
% - hard boundary -> plot feasibility over normalized lgradb
% - difficult constraint -> plot feasibility of model over feasibility of the optimization results

\subsection{Variable-shape baselines}
\label{sect:variable_shape_baselines}

The fixed-shape baselines specialize to a single topology. The variable-shape
track instead asks one model to cover the full range of field periods
$\nfp\in\{2,3,4,5\}$ and coil counts $\ncoils\in\{4,\dots,8\}$ in the training
distribution. We condition on the discrete topology by appending one-hot
encodings of $\nfp$ and $\ncoils$ to the input, and introduce two architectures
that emit a variable number of coils (full details in
\autoref{sect:appendix:model_config_and_architecture}). The \textbf{\maskedmlp{}} predicts
a fixed maximum of eight coil slots in one forward pass and masks the slots
beyond the requested $\ncoils$. The \textbf{\percoilmlp{}} instead applies one
weight-shared MLP to each coil slot, conditioned on the shared design context
plus the slot's one-hot index and normalized position
$(\text{slot}+\tfrac12)/\ncoils$, so it represents any coil count by
construction. As in the fixed-shape track we also
evaluate ensembles (\textbf{\ensemblemaskedmlp{}}, and \textbf{\ensemblepercoilmlp{}}) that
average $M{=}5$ members in coilset space.

\begin{wrapfigure}{r}{0.6\textwidth}
    \vspace{-0.5em}
    \centering
    \includegraphics[width=\linewidth]{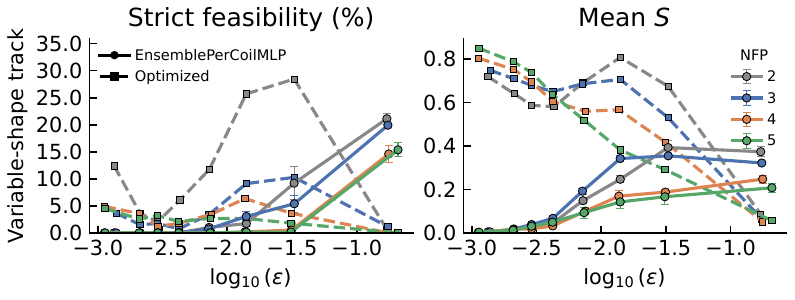}
    \caption{Strict feasibility (left) and mean score $S$ (right), grouped by reference mean normalized field error and field periods.
    Solid: EnsemblePerCoilMLP, mean $\pm$ one standard deviation across three training seeds.
    Dashed: DESC coilsets. DESC uses the three original geometric bounds; the model uses all four reference-achieved requirements.}
    \label{fig:feasibility_vs_field_error}
    \vspace{-1em}
\end{wrapfigure}

\autoref{fig:baselines_by_tier} (bottom) and \autoref{tab:baselines_variable}
report results on the $N{=}8{,}058$ variable-shape instances. The qualitative
picture matches the fixed-shape track: strict feasibility is $0\%$ on the tight
and medium tiers, with every feasible coilset in the loose tier. Ensembling
again lifts mean $\bar S$ substantially (\percoilmlp{}: $0.082\!\to\!0.137$),
and \ensemblepercoilmlp{} is the strongest variable-shape model on every
aggregate metric ($\bar S = 0.137$, strict $3.1\%$). Notably, evaluated on the
fixed-shape subset these topology-general models match or exceed the dedicated
fixed-shape baselines --- \ensemblepercoilmlp{} reaches $\bar S = 0.18$ at
$(\nfp{=}3,\ncoils{=}5)$ --- so sharing capacity across the full topology
distribution, in which $\nfp{=}3$ is oversampled, does not hurt and can help.
The per-coil ensemble is the model we carry forward as an optimization warm
start (\autoref{sect:optimization_warm_start}).

% \begin{table}[t]
% \tiny
% \centering
% \caption{Leaderboard on the variable-shape track ($N{=}8{,}058$ instances spanning $\nfp\in\{2,\dots,5\}$, $\ncoils\in\{4,\dots,8\}$), mean $\pm\,1\sigma$ over three training seeds, ranked by $\bar S$. Columns as in \autoref{tab:baselines}.}
% \label{tab:baselines_variable}
% \begin{tabular}{@{}lcccc@{}}
% \toprule
% Architecture & $\bar S$ & $\mathrm{med}\,S$ & Soft (\%) & Strict (\%) \\
% \midrule
% \ensemblepercoilmlp & $\mathbf{0.137\pm0.006}$ & $\mathbf{1.8\!\times\!10^{-2}}$ & $\mathbf{4.6\pm0.2}$ & $\mathbf{3.1\pm0.1}$ \\
% \ensemblemaskedmlp  & $0.128\pm0.005$          & $1.4\!\times\!10^{-2}$          & $4.3\pm0.3$          & $2.9\pm0.2$          \\
% \percoilmlp         & $0.082\pm0.005$          & $2.3\!\times\!10^{-3}$          & $3.6\pm0.1$          & $2.5\pm0.1$          \\
% \maskedmlp          & $0.081\pm0.005$          & $2.3\!\times\!10^{-3}$          & $3.3\pm0.4$          & $2.2\pm0.3$          \\
% \bottomrule
% \end{tabular}
% \end{table}

\paragraph{Field error remains the dominant failure mode.}
The reference coilset's mean normalized field error has the strongest correlation with both strict feasibility and score $S$ (\autoref{tab:feasibility_by_constraints}).
Requirements are derived from the reference coilsets' achieved metrics; the field-error requirement uses their maximum sampled normalized error.
Instances with larger reference mean field errors generally achieve higher feasibility and mean scores (\autoref{fig:feasibility_vs_field_error}).

% \begin{wrapfigure}{r}{0.6\linewidth[t]
% \centering
% \includegraphics[width=\linewidth]{figures/performance_vs_field_error.pdf}
% \caption{Strict-feasibility rate (left) and mean score $S$ (right)
% for EnsemblePerCoilMLP (solid) and DESC optimized coilsets (dashed) on the variable-shape evaluation set,
% grouped by requested normalized averaged field-error and number of field periods.
% In case of the EnsemblePerCoilMLP,
% values are the mean correlation $\pm$ one standard deviation across three independently trained seeds.
% }
% \label{fig:feasibility_vs_field_error}
% \end{wrapfigure}

\subsection{Learned coilsets as optimization warm starts}
\label{sect:optimization_warm_start}

\begin{wrapfigure}{r}{0.6\textwidth}
    \vspace{-1em}
    \centering
    \includegraphics[width=\linewidth]{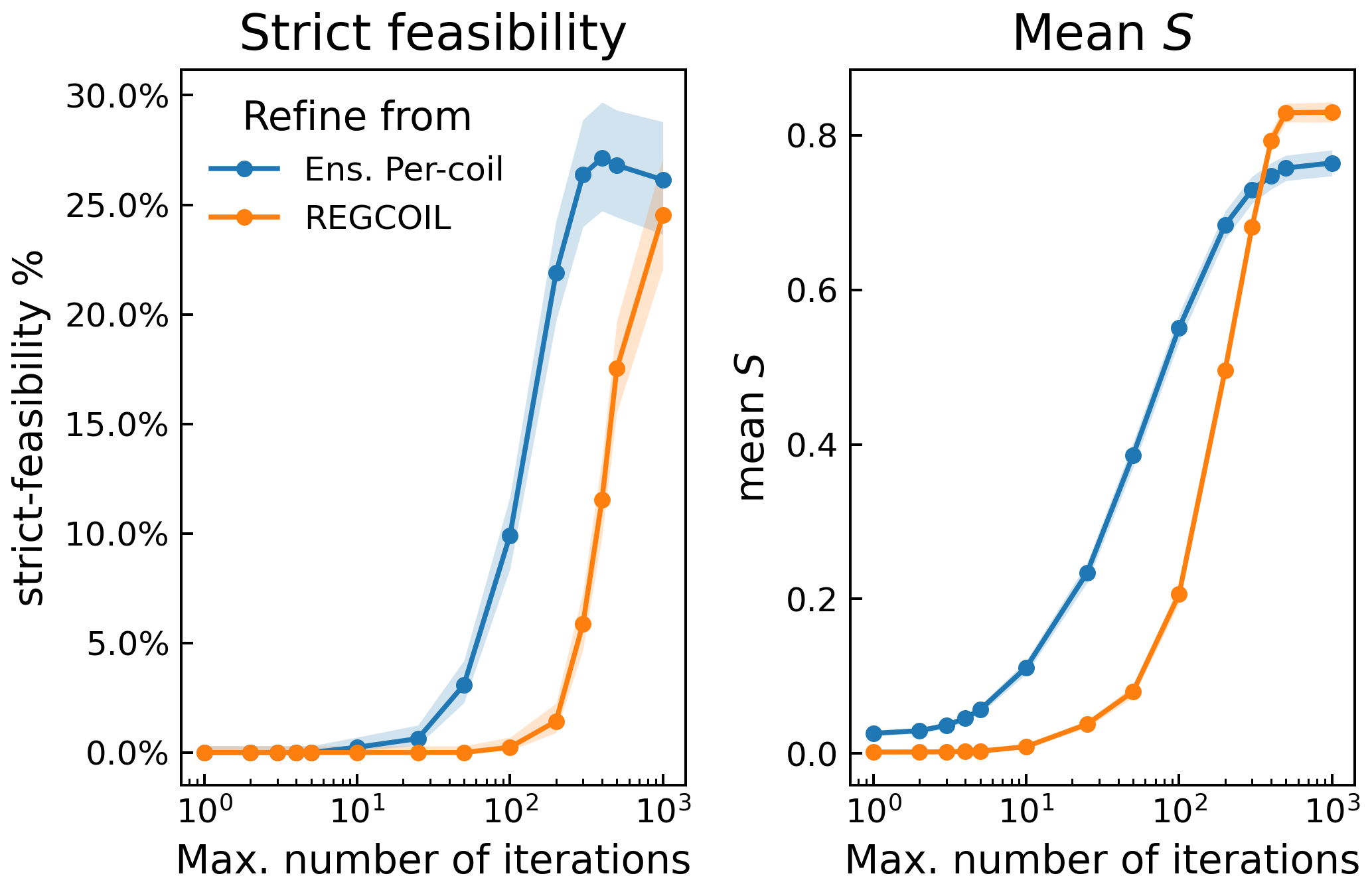}
    \caption{Strict-feasibility rate (left) and mean score $S$ (right) against the iteration budget, for the \regcoil and learned (per-coil ensemble) starts.
    Each instance uses its dataset metrics as targets; the instances comprise up to 100 per $(\nfp, \ncoils)$ pair.
    Markers show the tested budgets.
    Shaded bands are instance-level 95\% confidence intervals (Wilson for feasibility, Student-$t$ for $S$); they do not account for shared boundaries.}
    \label{fig:optimization_warm_start_aggregates}
    \vspace{-1em}
\end{wrapfigure}

A trained model can also initialize a conventional coil optimization.
On 1{,}289 evaluation instances from the tight stratum of the variable-shape evaluation split, we run the DESC-based optimization used to generate the dataset (\autoref{sect:dataset}) from two starts: the per-coil ensemble's single-pass prediction (\emph{learned start}) and the \regcoil coilset the pipeline normally uses (\emph{\regcoil start}).
We stop each run at a fixed iteration budget and score the result (\autoref{fig:optimization_warm_start_aggregates}).

Up to 300 iterations, the learned start achieves a higher mean score and wins on most instances.
In 100 iterations it beats the mean score of the \regcoil start after 200, and after 200 iterations it is strictly feasible more often than the \regcoil start after the dataset's full budget of 500 (22\% vs.\ 18\%).\footnote{The instance-level tests in \autoref{app:optimization_warm_start} do not account for shared plasma boundaries.}

The \regcoil start overtakes on mean score between 300 and 400 iterations and is ahead at 500 (0.83 vs.\ 0.76), but is still less often strictly feasible (18\% vs.\ 27\%).
By 1000 iterations the feasibility gap is no longer significant (26\% vs.\ 25\%, $p = 0.4$): the learned start levels off at about 26\% from 300 iterations on, whereas the \regcoil start keeps improving.
The per-requirement results explain this split (\autoref{app:optimization_warm_start}): the two starts have opposite weaknesses.
\regcoil coils are well separated but too close to the plasma, whereas learned coils more often clear the plasma but sit too close together.
Infeasible \regcoil-start runs tend to miss several requirements narrowly, which keeps their score high; infeasible learned-start runs more often miss one requirement by a wide margin.
The learned start is therefore most useful under a limited budget, such as inside an outer configuration-optimization loop or a large parameter scan.
Coil-to-coil spacing ($\minnormdcc$), the one requirement where the learned start still trails at 500 iterations, is the clearest target for improving the model.

\section{Discussion}
\label{sect:discussion}

% CoilStellaration complements existing equilibrium and coil databases by combining 177,000 optimized coilsets with varied engineering requirements and a standardized benchmark for conditional coil generation.
% The paired plasma boundaries, coil geometries, currents, and achieved metrics support studies of how plasma shape and engineering requirements affect coil design.
% Together with the released baselines and evaluation code, these resources provide a common starting point for machine learning and optimization researchers to develop and compare new methods.

% Magnetic-field accuracy remains the main obstacle to satisfying all benchmark requirements, motivating models and training objectives that account directly for the coil-generated field.
% Warm-start experiments also suggest that learned predictions can improve the early progress of conventional optimization, supporting further work on combining learning with numerical refinement.

% The present dataset uses filamentary coils and a limited set of engineering constraints, with evaluation on target boundaries rather than the resulting free-boundary equilibria.
% Within this scope, CoilStellaration supports methods that could reduce the cost of exploring stellarator coil designs.

CoilStellaration complements existing equilibrium and coil databases by combining 177,000 optimized coilsets with varied engineering requirements and a standardized benchmark for conditional coil generation.
Together with the released baselines and evaluation code, these resources provide a common starting point for machine learning and optimization researchers to develop and compare methods.
Magnetic-field accuracy remains the main obstacle to satisfying all requirements, motivating models and training objectives that account directly for the coil-generated field.
Warm-start experiments suggest that learned predictions can improve the early progress of conventional optimization, supporting further work on combining learning with numerical refinement.

The present dataset uses filamentary coils and a limited set of engineering constraints, with evaluation on target boundaries rather than the resulting free-boundary equilibria.
The main risk is therefore over-reliance on generated coilsets in a safety-critical domain: they are starting points for optimization and validation, not final designs.
Within this scope, CoilStellaration supports methods that could reduce the cost of exploring stellarator coil designs, accelerating progress toward fusion as a low-carbon energy source.

\begin{ack}
This work was independently funded by Proxima Fusion, and supported by the German Federal Ministry of Research, Technology and Space (BMFTR) under grant number 13F1012A (Joint Project: FUSKI). The authors are responsible for the content of this publication.
\end{ack}

\bibliographystyle{plain-six}
\bibliography{references, references_additional}

%%%%%%%%%%%%%%%%%%%%%%%%%%%%%%%%%%%%%%%%%%%%%%%%%%%%%%%%%%%%

\appendix

\section{Comparison with existing stellarator datasets}
\label{app:dataset_comparison}

\autoref{tab:dataset_comparison} compares CoilStellaration with
ConStellaration~\cite{cadenaConStellarationDatasetQIlike2025} and
QUASR~\cite{giulianiComprehensiveExplorationQuasisymmetric2025}.
ConStellaration provides QI-like plasma boundaries together with their
ideal-MHD equilibria and performance metrics. QUASR provides QA and QH
vacuum configurations with associated coilsets, obtained through direct
coil optimization. CoilStellaration uses plasma boundaries from
ConStellaration as fixed targets and optimizes multiple coilsets under
different engineering requirements.

\begin{table}[t]
    \centering
    \small
    \caption{
        Comparison of ConStellaration, QUASR, and CoilStellaration in terms of number of plasma boundaries, coilsets, and their properties.
        Engineering limits are prescribed requirements, not achieved
        metric ranges. 
        $a$ is the plasma minor radius,
        $A=R_0/a$ is the aspect ratio, and $b_{\mathrm{CP}}$,
        $b_{\mathrm{CC}}$, and $b_{\kappa}$ are sampled requirement
        thresholds.
        A dash indicates that a quantity is not applicable.
    }
    \label{tab:dataset_comparison}
    \begin{tabularx}{\textwidth}{
        @{}p{0.22\textwidth}
        *{3}{>{\raggedright\arraybackslash}X}@{}
    }
        \toprule
        Property
        & ConStellaration
        & QUASR
        & CoilStellaration \\
        \midrule

        Plasma boundaries
        & $\sim 158{,}000$
        & $\sim 370{,}000$
        & 14,496 target boundaries \\
        \addlinespace

        Coilsets
        & ---
        & $\sim 370{,}000$
        & 266,549 REGCOIL initializations;
          177,341 completed DESC coilsets \\
        \addlinespace

        Coilsets per boundary
        & ---
        & One paired coilset per entry
        & On average, 18.4 REGCOIL initializations
          and 12.2 DESC coilsets \\
        \addlinespace

        Configuration family
        & QI-like
        & QA and QH
        & QI-like \\
        \addlinespace

        Number of field periods
        & 1--5
        & 1--6
        & 2--5 \\
        \addlinespace
        
        Unique coils per half field period
        & ---
        & 1--13
        & 4--8 \\
        \addlinespace
        
        Optimization methodology
        & Stage-one plasma-boundary optimization through separate DESC
and VMEC++ pipelines
        & Single-stage direct coil optimization:
          near-axis initialization, BoozerLS, then BoozerExact
          refinement
        & Stage-two coil optimization:
          REGCOIL initialization followed by DESC refinement \\
        \addlinespace

        Optimization objectives
        & QI-field mismatch
        & QS error; additional Boozer-equation residual
          on BoozerLS surfaces
        & Weighted quadratic flux and coil length \\
        \addlinespace

        Prescribed plasma properties
        & Target QI field, aspect ratio, edge rotational
          transform, and elongation; mirror ratio encoded
          in the target field
        & Aspect ratio and mean rotational transform;
          surface shapes determined by the coils
        & Fixed target plasma boundary \\
        \addlinespace

        Coil length requirement
        & ---
        & $\sum_{i=1}^{N_C} L_i \leq L_{\max}$,
          with $L_{\max}=N_C L_{\mathrm{target}}$
          per half field period
        & No explicit length bound;
          coil length penalized in the objective \\
        \addlinespace
        
        Coil-to-plasma distance requirement
        & ---
        & ---
        & $d_{\mathrm{CP}}/a \geq b_{\mathrm{CP}}$,
          $b_{\mathrm{CP}}\in[0.5,\,1.0]$ \\
        \addlinespace

        Coil-to-coil distance requirement
        & ---
        & $d_{\mathrm{CC}}/a \geq 0.1A$
        & $d_{\mathrm{CC}}/a \geq b_{\mathrm{CC}}$,
          $b_{\mathrm{CC}}\in[0.3,\,0.7]$ \\
        \addlinespace

        Coil curvature requirement
        & ---
        & $a\kappa_{\max} \leq 5/A$
        & $a\kappa_{\max} \leq b_{\kappa}$,
          $b_{\kappa}\in[1.5,\,5.0]$ \\

        \bottomrule
    \end{tabularx}
\end{table}

\paragraph{Magnetic field accuracy.}
The QI and QS residuals measure departures from the desired magnetic
field properties,
whereas the weighted quadratic flux measures the tangency
of the coil-generated field to a given surface.
Therefore, these quantities assess different aspects of a design.
In CoilStellaration, the target boundary is prescribed independently
of the coils, so the weighted quadratic flux measures how closely the coilset reproduces that boundary.

\paragraph{Interpretation of engineering requirements.}
In QUASR, the coil-length budget is a prescribed design parameter,
while in CoilStellaration is treated as an objective to be minimized.
QUASR explores trade-offs among quasisymmetry, aspect ratio, rotational transform, and coil length while allowing the magnetic surfaces to change.
CoilStellaration instead varies engineering requirements for a prescribed target boundary.

\paragraph{Prescribed requirements and achieved feasibility.}
The requirement intervals in \autoref{tab:dataset_comparison}
describe optimization inputs, not achieved metric ranges.
QUASR discards designs whose prescribed physics or engineering
constraints remain violated beyond its stated tolerance after
penalty refinement.
For CoilStellaration, completion of the coil optimization does
not imply that every sampled requirement has been satisfied,
and the dataset quality filters are distinct from a test of
compliance with all prescribed requirements.
Achieved metrics must therefore be evaluated against the
requirements of each sample when assessing feasibility.

\section{Details regarding the optimization warm-start}
\label{app:optimization_warm_start}

The 1{,}289 instances cover 308 plasma boundaries, each paired with several requirement vectors.
All runs use the dataset pipeline's optimizer (\autoref{sect:dataset}) with the instance's DESC equilibrium from the dataset, and each budget is a separate run from the initial coilset.
DESC refinement uses the original sampled geometric bounds. Evaluation compares achieved metrics with the reference coilset's achieved metrics.
Runs that fail or time out are excluded: none up to 300 iterations, about 1\% at 400 and 500, and 7\% at 1000.
Paired tests use the instances with results from both starts: the Wilcoxon signed-rank test for scores and the exact McNemar test for pass/fail outcomes.
These instance-level intervals and tests do not account for dependence among requirement vectors sharing a boundary; they do not quantify uncertainty across new plasma boundaries.
The learned start scores higher on 71--86\% of instances at every budget up to 200 iterations ($p < 10^{-50}$) and on 61\% at 300 ($p = 1 \times 10^{-9}$).
At 500 iterations, the \regcoil start scores higher ($p = 2 \times 10^{-9}$) but is less often strictly feasible ($p = 1 \times 10^{-9}$).
The cross-budget comparisons in \autoref{sect:optimization_warm_start} are also significant: $p = 5 \times 10^{-7}$ for mean score (learned at 100 vs.\ \regcoil at 200) and $p = 0.004$ for strict feasibility (learned at 200 vs.\ \regcoil at 500).
At 1000 iterations, the strict-feasibility gap is no longer significant ($p = 0.39$).

\begin{figure}[htb]
    \vspace{-1em}
    \centering
    \includegraphics[width=\linewidth]{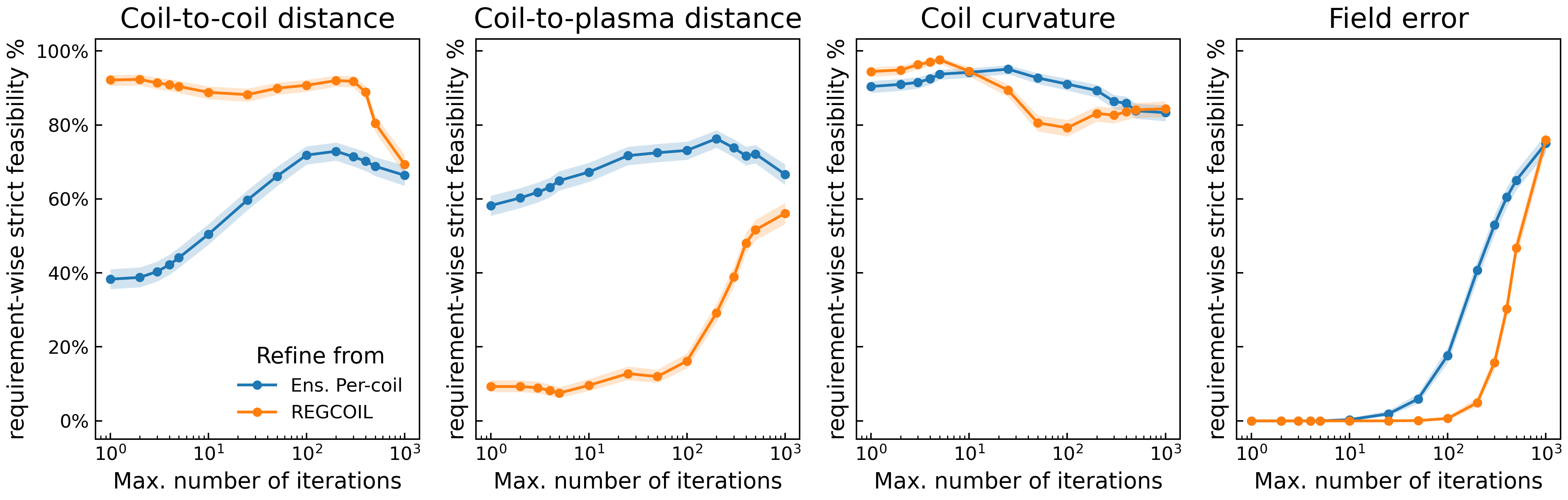}
    \caption{Fraction of runs meeting each requirement against the iteration budget, for the learned and \regcoil starts.
    Shaded bands are 95\% Wilson intervals over the instances with results from both starts.}
    \label{fig:warm_start_requirements}
    \vspace{-1em}
\end{figure}

\autoref{fig:warm_start_requirements} and \autoref{tab:warm_start_requirements} break the results down by requirement.
Initially, the \regcoil start meets $\minnormdcc$ on 92\% of instances but $\minnormdcp$ on only 10\%; the learned start shows the opposite pattern (38\% and 58\%).
Up to 200 iterations, the learned start improves much faster on $\maxfielderror$ and stays ahead on $\minnormdcp$, which outweighs its $\minnormdcc$ deficit.
From 200 to 500 iterations, $\maxfielderror$ keeps improving for both starts while $\minnormdcc$ deteriorates, most for the \regcoil start (91\% to 80\%), consistent with the optimizer trading coil spacing for field accuracy.
At 1000 iterations, the two starts meet $\maxfielderror$ and $\maxnormcurvature$ equally often; the learned start still meets $\minnormdcp$ more often (67\% vs.\ 56\%, $p = 7 \times 10^{-9}$), and the \regcoil start's lead on $\minnormdcc$ shrinks to 69\% vs.\ 66\% ($p = 0.009$).
Of the runs still infeasible at 500 iterations, learned-start runs more often violate exactly one requirement (61\% vs.\ 47\%), but by a wider margin: the median soft score of their worst requirement is 0.39, compared with 0.65 for \regcoil-initialized runs.

\begin{table}[htb]
  \centering
  \caption{Fraction of runs meeting each requirement, and all of them (strict feasibility), after 1 and 500 iterations, over the instances with results from both starts.
  At 500 iterations, every difference except curvature is significant (exact McNemar test, $p < 10^{-8}$; curvature $p = 0.88$).}
  \label{tab:warm_start_requirements}
  \begin{tabular}{@{}lcccc@{}}
    \toprule
    & \multicolumn{2}{c}{1 iteration} & \multicolumn{2}{c}{500 iterations} \\
    \cmidrule(lr){2-3} \cmidrule(lr){4-5}
    Requirement & Per-coil ensemble & REGCOIL & Per-coil ensemble & REGCOIL \\
    \midrule
    $\minnormdcc$       & 38\% & 92\% & 69\% & 80\% \\
    $\minnormdcp$       & 58\% & 10\% & 72\% & 52\% \\
    $\maxnormcurvature$ & 90\% & 94\% & 84\% & 84\% \\
    $\maxfielderror$    & 0\%  & 0\%  & 65\% & 47\% \\
    \midrule
    All                 & 0\%  & 0\%  & 27\% & 18\% \\
    \bottomrule
  \end{tabular}
\end{table}

\section{Detailed Filtering Pipeline for ConStellaration Boundaries}
\label{sect:appendix:boundary_filtering_details}

This section details the explicit filtering sequence applied to map candidate boundaries to solved equilibria, which subsequently act as the geometric targets for our coilset dataset, as touched upon in \autoref{sect:dataset}.
The selection pipeline acts as a sequential filter $F_1 \circ F_2 \circ \dots \circ F_k$, applied to the superset of generated configurations.

\subsection{Optimization and Convergence Constraints}

Candidate boundaries are generated via a multi-stage pipeline consisting of initialization, boundary optimization (via VMEC++ or standard VMEC), and forward-model metric evaluation.
We define a Boolean failure vector for each configuration $i$, $\mathbf{e}^{(i)} \in \{0, 1\}^5$, representing the presence of errors across these sub-routines.
The first filter, $F_1$, removes any configuration where upstream generation failed:
\begin{equation}
    F_1(x) = \begin{cases} 
      \text{Keep}, & \text{if } \sum_{j} \mathbf{e}^{(i)}_j = 0 \\
      \text{Drop}, & \text{otherwise}
   \end{cases}
\end{equation}

Furthermore, to guarantee that the boundary geometry is the result of active optimization of omnigenity rather than a warm-start initialization, the filter $F_2$ restricts the generation method.
If $\mathcal{M}$ represents the set of applied boundary generation methods, we require $\mathcal{M} \in \{\text{VMEC++}, \text{VMEC}\}$. Boundaries generated exclusively via quasi-poloidal or near-axis expansion analytical formulations are excluded, as their QI metrics are not representative of optimal targets for coil design.

\subsection{Physical and Metric Validations}

Stellarator boundary orientation dictates the sign of the rotational transform.
For consistency across downstream coil parameterization tasks, filter $F_3$ strictly enforces the standard positive-$\iota$ orientation.
Let $N_{\mathrm{FP}}$ be the number of field periods.
The edge rotational transform is recovered and constrained via:
\begin{equation}
    \iota_{\mathrm{edge}} = \left( \frac{\iota_{\mathrm{edge}}}{N_{\mathrm{FP}}} \right) N_{\mathrm{FP}} > 0.
\end{equation}

Filter $F_4$ subsequently evaluates the availability of the quasi-isodynamic residual, $\mathcal{Q}$. Configurations are discarded if $\mathcal{Q} = \emptyset$, which typically occurs when the Boozer coordinate transformation or the turbulent metric computations diverge despite successful VMEC++ convergence.

\subsection{Rotational Symmetry and Field Period Constraints}

Stellarator geometries are fundamentally characterized by their discrete toroidal symmetry, defined by the number of field periods, $N_{\mathrm{FP}}$. Downstream coil optimization tasks often target specific symmetry classes to satisfy engineering constraints regarding modularity and maintenance access.
Consequently, a final optional, yet highly relevant, filter ($F_5$) restricts the dataset to a targeted subset of field periods, $\mathcal{S}_{\mathrm{NFP}}$.

We restrict this release to $N_{\mathrm{FP}}\in\{2,3,4,5\}$.
Therefore, the field period filter acts to enforce the allowable topological bounds:
\begin{equation}
    F_5(x) = \begin{cases} 
      \text{Keep}, & \text{if } N_{\mathrm{FP}}^{(i)} \in \mathcal{S}_{\mathrm{NFP}} \cap \{2, 3, 4, 5\} \\
      \text{Drop}, & \text{otherwise}
   \end{cases}
\end{equation}
While $\mathcal{S}_{\mathrm{NFP}}$ typically defaults to $3$ for baseline analyses, it can be extended to cover the full available interval $[2, 5]$.
% Any subset query attempting to sample configurations with $N_{\mathrm{FP}} = 5$ will systematically yield an empty set, ensuring that downstream coil generative models are not trained on sparse or unsupported symmetry spaces.

\subsection{Equilibrium Cross-Referencing}

Each retained boundary is linked to a solved VMEC++ equilibrium with recorded force-balance diagnostics.
We execute an inner join between the filtered boundaries and a precomputed library of VMEC++ solves.

Let $\hat{F}_{\mathrm{mean}}$ and $\hat{F}_{\mathrm{max}}$ denote the normalized mean and maximum absolute force errors computed over the computational flux-surface grid of the VMEC++ solve.
A boundary is only retained if it maps to a successfully instantiated and solved VMEC++ equilibrium where these final objective metrics are quantified.
\autoref{tab:filter_summary} summarizes the complete filtering hierarchy.

\begin{table}[h]
    \centering
    \caption{Summary of the Sequential Target Boundary Filtering Pipeline}
    \label{tab:filter_summary}
    \begin{tabular}{@{}llp{8cm}@{}}
        \toprule
        \textbf{Stage} & \textbf{Criterion} & \textbf{Physical / Numerical Rationale} \\
        \midrule
        $F_1$ & Upstream Convergence & Discard configurations with missing attributes caused by non-convergence in stage-1 (optimization) or stage-2 (forward models). \\
        $F_2$ & Optimization Provenance & Require the boundary to originate from active VMEC++ or standard VMEC omnigenity optimization, rejecting unoptimized initial guesses.
\\
        $F_3$ & Positive Orientation & Enforce $\iota_{\mathrm{edge}} > 0$ to standardize spatial orientation for downstream control-coil synthesis and neural network ingestion.
\\
        $F_4$ & Metric Validity & Require a finite Goodman QI residual ($\mathcal{Q} \neq \emptyset$). \\
        $F_5$ & Field Period Bounds & Restrict this release to $N_{\mathrm{FP}}\in\{2,3,4,5\}$. \\
        $F_6$ & Equilibrium Mapping & Exclude boundaries that fail to instantiate or converge to a force-balanced state within the VMEC++ computational framework.
\\
        \bottomrule
    \end{tabular}
\end{table}

MHD stability and engineering feasibility require separate analyses.

\section{Dataset breakdown}

\subsection{Distribution ranges of the engineering requirements for the sampling pipeline}
\label{sect:appendix:ranges_engineering_requirements}
In the sampling pipeline, we sample engineering requirements ($\minnormdcc$, $\minnormdcp$, and $\maxnormcurvature$) in ranges (\autoref{tab:reference_stellarators}) informed by relevant stellarator designs, either built, like the W7-X~\cite{beidlerPhysicsEngineeringDesign1990} and HSX~\cite{wiedmanCoilOptimizationQuasihelically2023} experiments, or proposed, like the NCSX~\cite{nelson2003design} experiment and the power plant reactors Stellaris~\cite{lionStellarisHighfieldQuasiisodynamic2025} and Infinity Two~\cite{hegna2025infinity}.
% See table:
% https://docs.google.com/document/d/1Jr2YWs7u_r8Waz5jggJIwvgNE-IJHY_37QMAvphaRlQ/edit?tab=t.y727vyu5tay1
\begin{table}[]
\caption{Reference stellarators from which the requirements sampling range was informed.}
\label{tab:reference_stellarators}
\begin{tabular}{@{}lccc@{}}
\toprule
Reference Design & $\minnormdcc$ (coil-to-coil distance) & $\minnormdcp$ (coil-to-plasma distance) & $\maxnormcurvature$ (coil curvature)\\
\midrule
W7-X~\cite{beidlerPhysicsEngineeringDesign1990} & 0.53 & 0.70 & 1.31 \\
HSX~\cite{wiedmanCoilOptimizationQuasihelically2023} & 0.76 & 1.18 & 1.4 \\
NCSX~\cite{nelson2003design} & 0.38 & 0.67 & 3.18 \\
Stellaris~\cite{lionStellarisHighfieldQuasiisodynamic2025} & 0.51 & 1.05 & 2.03 \\
Infinity Two~\cite{hegna2025infinity} & 0.52 & 0.97 & 2.4 \\
Sampled Range & 0.3 - 0.7 & 0.5 - 1.0 & 1.5 - 5.0 \\
\bottomrule
\end{tabular}
\end{table}

\subsection{Dataset composition}
\label{sect:appendix:detailed_dataset_breakdown}
Coilset are sampled for boundaries across different values of $\nfp$.
\autoref{tab:appendinx/dataset-composition} reports the breakdown of the number of \regcoil and \desc coilsets per $\nfp$ category, and their relative fractions.
Noteworthy, the $\nfp=3$ category has been oversampled in order to use it for machine learning purposes.
\begin{table}[htbp]
  \centering
  \caption{Dataset composition by number of field periods ($\nfp$).
           \desc{}\,\%\,of\,\desc{} is the fraction of each
           $\nfp$ group within all \desc{} coilsets.
           \desc{}/\regcoil{} is the completion rate of the
           \desc{} optimization per $\nfp$.}
  \label{tab:appendinx/dataset-composition}
  \begin{tabular}{r rr rr r}
    \toprule
    $\nfp$
      & \regcoil{} & (\%)
      & \desc{} & \% of \desc{}
      & \desc{}/\regcoil{} (\%) \\
    \midrule
    2 & 30\,481 & 11.4 & 30\,234 & 17.0 & 99.2 \\
    3 & 104\,075 & 39.0 & 89\,503 & 50.5 & 86.0 \\
    4 & 63\,609 & 23.9 & 33\,077 & 18.7 & 52.0 \\
    5 & 68\,384 & 25.7 & 24\,527 & 13.8 & 35.9 \\
    \midrule
    Total & 266\,549 & 100.0 & 177\,341 & 100.0 & 66.5 \\
    \bottomrule
  \end{tabular}
\end{table}

\subsection{Breakdown of the filter}
\label{sect:appendix:filter_breakdown}
We apply a quality filter to remove coilsets with large discrepancies between
the sampled engineering requirements and the measured values. With relative
error defined as
$e_x=(x_{\mathrm{measured}}-x_{\mathrm{target}})/x_{\mathrm{target}}$,
we require $e_{\mathrm{CP}}>-0.5$, $e_{\mathrm{CC}}>-0.5$, and $e_\kappa<1.0$.
These cuts remove large geometric violations; no field-error cut is applied.
The reference population comprises the 266,549 released samples with a
\regcoil{} initialization. \autoref{tab:selection-cuts} separates
\desc{} output availability, metric availability, and quality selection.
The mere fact that the output is missing does not establish an optimization failure mechanism.

\begin{table}[htbp]
\centering\small
\caption{Cumulative selection cut flow, overall and by field periods. Retention is relative to the released samples in each column at the time of submission.}
\label{tab:selection-cuts}
\setlength{\tabcolsep}{3pt}
\begin{tabular}{lrrrrr}
\toprule
Stage & All & $\nfp=2$ & $\nfp=3$ & $\nfp=4$ & $\nfp=5$ \\
\midrule
Released samples & 266\,549 & 30\,481 & 104\,075 & 63\,609 & 68\,384 \\
Successful optimization & 177\,341 & 30\,234 & 89\,503 & 33\,077 & 24\,527 \\
Full pipeline executed & 163\,908 & 30\,221 & 78\,906 & 33\,020 & 21\,761 \\
+ Coil-plasma error $> -0.5$ & 140\,965 & 26\,218 & 66\,237 & 29\,368 & 19\,142 \\
+ Coil-coil error $> -0.5$ & 138\,954 & 25\,692 & 65\,554 & 28\,949 & 18\,759 \\
+ Curvature error $< 1.0$ & 138\,765 & 25\,614 & 65\,455 & 28\,939 & 18\,757 \\
\midrule
Retention (\%) & 52.1 & 84.0 & 62.9 & 45.5 & 27.4 \\
\bottomrule
\end{tabular}
\end{table}

\begin{figure}[htbp]
\centering
\includegraphics[width=\textwidth]{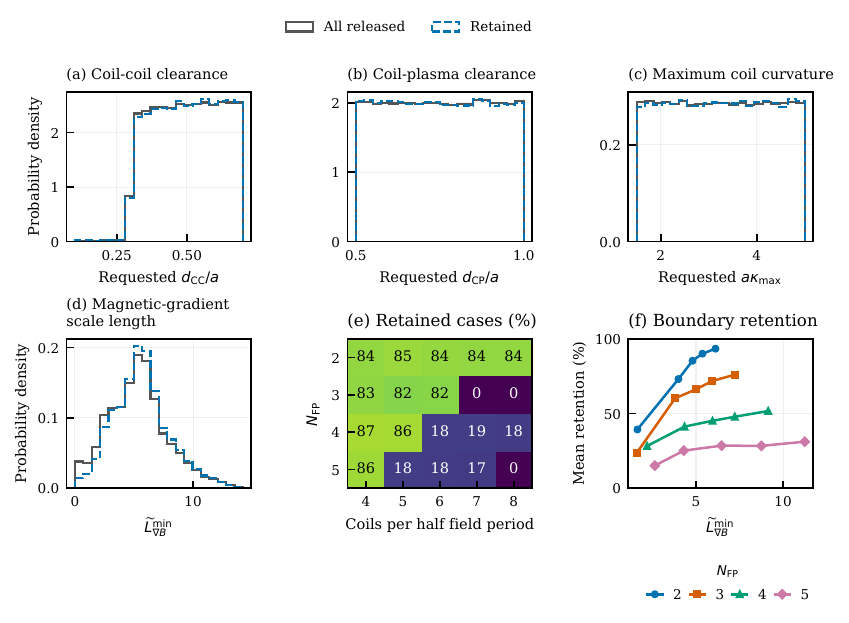}
\caption{Selection effects. (a--c) Probability densities of requested
coil--coil clearance, coil--plasma clearance, and maximum coil curvature
for released and retained samples. Maximum absolute differences between
the corresponding empirical cumulative distribution functions (CDFs) are
0.61, 0.38, and 0.40 percentage points,
respectively. (d) Probability density of the minimum normalized
magnetic-gradient scale length over unique boundaries; a retained boundary
has at least one retained coilset. Panels (a--d) use 20 shared equal-width
bins per panel and unit-area normalization for each population. (e) Retained fraction
of released samples for each $\nfp$ and coil count. (f) Mean per-boundary
retention in five scale-length quantile bins per $\nfp$, with equal weight
per boundary and points at the mean scale length of each bin.}
\label{fig:selection-bias}
\end{figure}

The retained fraction decreases from 84.0\% at $\nfp=2$ to 27.4\% at
$\nfp=5$. However, retention also depends strongly on coil count
(\autoref{fig:selection-bias}e): with four coils per half field period,
it ranges from 83.3\% to 87.0\% across $\nfp$, while some
other combinations have no retained samples. The aggregate trend therefore
cannot be attributed to $\nfp$ alone.

\autoref{fig:selection-bias}a--d compares the input distributions before
and after selection. The marginal distributions of the three geometric
requirements change little: their maximum cumulative-distribution differences
are at most 0.61 percentage points. For the minimum normalized
magnetic-gradient scale length $\widetilde{L}_{\nabla B}^{\min}$ --- a common proxy for expected coilset simplicity \cite{cadenaConStellarationDatasetQIlike2025,kappelMagneticGradientScale2024} --- we count
each boundary once. Of 14\,496 target boundaries,
13\,037 (89.9\%) retain
at least one coilset.

Retention also varies with coil count within each $\nfp$ group (\autoref{fig:selection-bias}e), indicating that the aggregate differences across $\nfp$ depend on the sampled coil-count distribution.

Finally, we define each boundary's retention rate as its number of retained
coilsets divided by its number of released samples, including zero-retention
boundaries. \autoref{fig:selection-bias}f averages these rates within
magnetic-gradient scale length quintiles, separately for each $\nfp$. Retention generally increases with
magnetic-gradient scale length; at $\nfp=2$, the lowest and highest quintiles have mean rates
of 39.3\% and 93.4\%, respectively. Thus, boundaries with more rapidly
varying fields are less completely represented.
This is expected as a larger magnetic-gradient scale length is associated with simpler or more engineerable coilsets.

\subsection{Objective normalization and quadrature}
\label{sect:appendix:objective_normalization}
\autoref{eq:coil-optimization-problem} expresses DESC's least-squares objective as a weighted sum of scalar length and normal-field penalties.
The implementation multiplies both residual vectors by $0.1$, so its scalar objective is $0.01\mathcal{J}$; this common factor does not change the exact constrained minimizer.
Coil lengths are divided by $\ell_*=(1/N_c)\sum_i L_i^0/(2\pi)$, computed from the initial coils.
For $N_s$ surface nodes, let $g_j$ denote the weights returned by DESC's surface grid (\texttt{grid.weights}) and $G_j=|\partial_\theta\mathbf{x}\times\partial_\zeta\mathbf{x}|$ the surface Jacobian.
DESC's normal-field residual at node $j$ is $\sqrt{N_s}\,g_j(\mathbf{B}_j\cdot\mathbf{n}_j)\sqrt{G_j}/(B_*R_*a_*)$, giving the effective surface weights $W_j=N_s g_j^2G_j$ in \autoref{eq:coil-optimization-problem}.
The curvature weights $w_j$ are coil-parameter grid spacings, without an arclength Jacobian.
The fixed normalization scales are computed when the objective is built:
$R_*$ is the boundary's zero-mode radial coefficient,
$a_*=\sqrt{|R_*^{\mathrm{nz}}Z_*^{\mathrm{nz}}|}$ uses the largest-magnitude nonzero radial and vertical surface coefficients, and
$B_*=1.25|\Psi|/(\pi a_*^2)$, with $\Psi$ the equilibrium toroidal flux.
The length scale $\ell_*$ and the DESC scale $a_*$ are distinct from the plasma minor radius used to normalize the reported geometric metrics.

\section{Dataset Metrics Specification}
\label{sect:appendix:dataset_metrics}

This section details the physical, geometric, and magnetic metrics calculated for each coilset in the dataset. To facilitate scale-invariant evaluations, many geometric quantities are normalized by the plasma minor radius. Metrics are evaluated on fixed surface and coil grids to ensure symmetry and numerical consistency across the dataset.

\subsection{Core Geometric and Configuration Metrics}
These fundamental metrics characterize the physical layout and basic constraints of the coil configurations.

\begin{itemize}
    \item \textbf{Minor Radius:} The plasma minor radius $a$ (meters), utilized as the primary normalization scale for lengths, distances, curvatures, and torsions across the dataset.
    \item \textbf{Coil Currents:} The assigned electrical current for each structurally unique coil (amperes).
    \item \textbf{Coil Lengths:} The total arc length of each coil (meters). The derived normalized length $\normlength$ is calculated as the absolute length divided by $a$. Any separately rescaled plot quantity is denoted $L/(2\pi a)$.
    \item \textbf{Coil Curvatures:} The pointwise local curvature along the coil paths ($\mathrm{m}^{-1}$). The derived normalized curvature is scaled by multiplying by $a$.
    \item \textbf{Coil-to-Coil Minimum Distance:} The minimum Euclidean distance between any two coils in the fully symmetry-expanded configuration (meters). Normalized distance is divided by $a$.
    \item \textbf{Coil-to-Plasma Minimum Distance:} The minimum Euclidean distance from the coil set to the targeted plasma boundary surface (meters). Normalized distance is divided by $a$.
\end{itemize}

\subsection{Surface Magnetic Field Metrics}
These metrics evaluate the accuracy of the magnetic confinement, evaluated over the target plasma boundary. Let $B$ denote the magnetic field generated by the coils, $B_n$ represent the component of $B$ normal to the surface, and $dS$ denote the local surface area element.

\begin{itemize}
    \item \textbf{Normalized Field Error:} The local error is $e_j=|B_j\cdot n_j|/\|B_j\|$. The surface mean is $\epsilon_{\mathrm{mean}}=\sum_j w_j e_j/\sum_j w_j$, with $w_j$ the surface-area quadrature weights. The maximum sampled error is $\epsilon_{\max}=\max_j e_j$; it sets the benchmark field-error requirement.
    \item \textbf{Local Quadratic Flux:} The surface-area-weighted normalized quadratic-flux density, calculated as $\frac{B_n^2}{|B|^2} dS$ ($\mathrm{m}^2$).
    \item \textbf{Quadratic Flux Residual:} The per-surface-node residual flux value, defined as $B_n \sqrt{dS}$ ($\mathrm{T} \cdot \mathrm{m}$).
\end{itemize}

\subsection{Diagnostic Metrics}
Computed alongside the core metrics, these diagnostics provide extended physical insights regarding the engineering feasibility and topological complexity of the configurations.

\begin{itemize}
    \item \textbf{On-Axis Average Magnetic Field:} The mean magnetic-field magnitude evaluated strictly along the magnetic axis over one field period (tesla).
    \item \textbf{Coil Linking Numbers:} A topological diagnostic representing the sum of absolute linking numbers between coils (dimensionless).
    \item \textbf{Coil Current-Lengths:} The product of current and arc length for each unique coil ($\mathrm{A} \cdot \mathrm{m}$), serving as a standard proxy for superconductor material costs.
    \item \textbf{Coil Integrated Curvatures:} The line integral of the curvature over the entirety of the coil curve (dimensionless; evaluates to $2\pi$ for strictly convex planar coils).
    \item \textbf{Coil Torsions:} The pointwise torsion along the coil curve ($\mathrm{m}^{-1}$). Normalized torsion is scaled by multiplying by $a$.
    \item \textbf{Coil Arclength Variances:} A quantitative measure of the nonuniformity in the parameterization of the generated coil curves.
    \item \textbf{Toroidal Flux:} The total toroidal magnetic flux passing through the plasma surface generated exclusively by the external coil field (webers).
    \item \textbf{Linking Current Consistency:} A consistency measure comparing the external coil linking currents directly to the targeted internal equilibrium currents (amperes).
\end{itemize}

\subsection{Baseline Requirement Inputs}
The baselines take four requirements: minimum normalized coil--coil distance, minimum normalized coil--plasma distance, maximum normalized curvature, and maximum normalized field error.
On-axis field and toroidal flux are diagnostic quantities, not inputs to these baselines.
The four inputs use the robust standardization in \autoref{app:input_representation_and_standardization}.

\section{Benchmark scoring details}
\label{sect:appendix:benchmark_scoring}

\subsection{Per-metric tolerances}
\label{sect:appendix:tolerances}
The per-metric tolerances $\alpha_i$ used in the soft-feasibility score
(\autoref{eq:soft-score}) are listed in \autoref{tab:appendix:tolerances}.
The tighter tolerance on the geometric metrics reflects the engineering
precision they require; field error has a looser tolerance because its
absolute scale is small in the dataset. We commit to these values once and
freeze them before any baseline run --- they are physics judgments tied to
the precision the dataset achieves on each metric, not hyperparameters.

\begin{table}[h]
\centering
\caption{Per-metric tolerances $\alpha_i$ in the soft-feasibility score
(\autoref{eq:soft-score}). A relative violation of $\alpha_i$ reduces $s_i$
to $1/e \approx 0.37$; a $2\alpha_i$ violation reduces it to $\approx 0.14$.}
\label{tab:appendix:tolerances}
\begin{tabular}{@{}lcc@{}}
\toprule
Metric & Direction & $\alpha_i$ \\
\midrule
$\minnormdcc$ (coil-to-coil distance)   & lower bound & $0.10$ \\
$\minnormdcp$ (coil-to-plasma distance) & lower bound & $0.10$ \\
$\maxnormcurvature$ (coil curvature)    & upper bound & $0.10$ \\
$\maxfielderror$ (field error)          & upper bound & $0.25$ \\
\bottomrule
\end{tabular}
\end{table}

\subsection{Per-metric breakdown of the baseline failure mode}
\label{sect:appendix:per-metric}
\autoref{tab:appendix:per-metric} reports the median per-metric soft score and
median violation $v$, pooled over the three training seeds per architecture, on
the fixed-shape track ($N{=}2{,}771$). Field error is the dominant failure across
all architectures ($s_\epsilon \sim 10^{-5}\text{--}10^{-4}$, median normalized violations $2.1$--$2.9$). Both ensembles have median soft scores near one for coil--plasma distance and curvature, while every architecture has a positive median coil--coil spacing violation. The four-fold geometric mean is dominated by the
smallest term, consistently $s_\epsilon$.

\begin{table}[h]
\centering
\caption{Median per-metric soft score and (median violation $v$) on the
fixed-shape track, pooled over the three training seeds per architecture. Soft
scores are bounded $[0,1]$ with $s{=}1$ at and beyond the threshold and
$s{=}1/e$ at $v{=}\alpha$ (see \autoref{eq:soft-score}).}
\label{tab:appendix:per-metric}
\begin{tabular}{@{}lcccc@{}}
\toprule
Architecture
  & $s_{d_{cc}}$
  & $s_{d_{cp}}$
  & $s_\kappa$
  & $s_\epsilon$ \\
\midrule
\mlp            & $0.02\;(v{=}0.39)$ & $0.35\;(v{=}0.11)$ & $0.14\;(v{=}0.19)$           & $1{\times}10^{-5}\;(v{=}2.8)$ \\
\resmlp         & $0.03\;(v{=}0.34)$ & $0.36\;(v{=}0.10)$ & $4{\times}10^{-3}\;(v{=}0.56)$ & $1{\times}10^{-5}\;(v{=}2.9)$ \\
\ensemblemlp    & $0.12\;(v{=}0.21)$ & $1.00\;(v{=}0)$    & $1.00\;(v{=}0)$              & $1{\times}10^{-4}\;(v{=}2.2)$ \\
\ensembleresmlp & $0.09\;(v{=}0.25)$ & $1.00\;(v{=}0)$    & $1.00\;(v{=}0)$              & $2{\times}10^{-4}\;(v{=}2.1)$ \\
\bottomrule
\end{tabular}
\end{table}

\begin{table}[h]
\centering
\caption{Spearman correlations between reference-coilset metrics and prediction performance
for EnsemblePerCoilMLP on the variable-shape evaluation set.
Values are the mean correlation $\pm$ one standard deviation across three independently trained seeds.
}
\label{tab:feasibility_by_constraints}
\begin{tabular}{lcc}
\hline
Reference metric & $\rho$ with strict feasibility & $\rho$ with score $S$ \\
\hline
$\widetilde d_{\mathrm{CC}}^{\min}$
    & $-0.116 \pm 0.008$ & $+0.218 \pm 0.005$ \\
$\widetilde d_{\mathrm{CP}}^{\min}$
    & $-0.214 \pm 0.004$ & $-0.073 \pm 0.008$ \\
$\widetilde\kappa_{\max}$
    & $+0.188 \pm 0.003$ & $-0.08 \pm 0.02$ \\
$\log_{10}(\text{reference mean normalized field error})$
    & $+0.259 \pm 0.004$ & $+0.42 \pm 0.03$ \\
\hline
\end{tabular}
\end{table}

\subsection{Sensitivity to the soft-score tolerance}
\label{sect:appendix:alpha_sensitivity}
The soft-feasibility score (\autoref{eq:soft-score}) depends on the per-metric
tolerances $\alpha_i$ (\autoref{tab:appendix:tolerances}), which we freeze before any
baseline run. Since the per-instance violations $v_i$ are stored, the score
$S = \exp\!\bigl(-\tfrac14\sum_i v_i/\alpha_i\bigr)$ can be recomputed for any
$\alpha$ without re-evaluating a model, so we can check whether the leaderboard
ordering depends on this choice. Strict feasibility ($v_i \le \num{1e-3}$) is
independent of $\alpha$ and is unaffected.

\begin{figure}[h]
  \centering
  \includegraphics[width=0.6\linewidth]{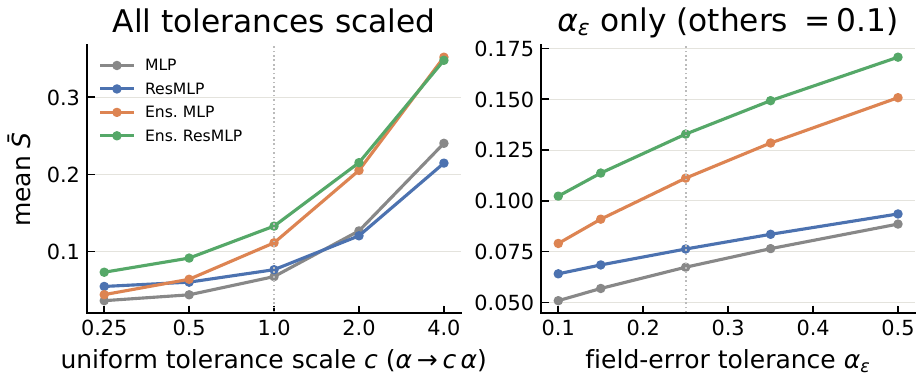}
  \caption{Mean score $\bar S$ on the fixed-shape track as the soft-score tolerance is
  varied: (left) all four tolerances scaled by a common factor $c$ (log axis; dotted
  line is the nominal $c{=}1$), and (right) only the field-error tolerance
  $\alpha_\epsilon$ (dotted line is the nominal $0.25$). Each point is the mean over
  three seeds. Field-error-only changes preserve the ranking; uniform scaling can change it.}
  \label{fig:alpha_sensitivity}
\end{figure}

\autoref{fig:alpha_sensitivity} shows the mean $\bar S$ on the fixed-shape track as we
(left) scale all four tolerances by a common factor $c$ and (right) vary only the
field-error tolerance $\alpha_\epsilon$, the dominant and most scrutinized term.
Varying only the field-error tolerance from $0.10$ to $0.50$ leaves the ordering unchanged.
Uniform scaling changes the ordering at $c = 0.25$, $2$, and $4$ (\autoref{tab:alpha_ranking}).

\begin{table}[h]
\centering
\caption{Kendall rank correlation $\tau$ of the fixed-shape mean-$\bar S$ ranking
against the nominal-$\alpha$ ranking, as the tolerances are varied. $\tau{=}1$ means
the four-architecture order is unchanged.}
\label{tab:alpha_ranking}
\begin{tabular}{@{}lccccc@{}}
\toprule
Uniform scale $c$        & $0.25$ & $0.5$  & $1$   & $2$    & $4$    \\
\midrule
$\tau$ vs nominal        & $0.67$ & $1.00$ & ---   & $0.67$ & $0.33$ \\
\midrule
Field-error $\alpha_\epsilon$ & $0.10$ & $0.15$ & $0.25$ & $0.35$ & $0.50$ \\
\midrule
$\tau$ vs nominal        & $1.00$ & $1.00$ & ---   & $1.00$ & $1.00$ \\
\bottomrule
\end{tabular}
\end{table}

\section{Detailed Model Configurations and Architectures}
\label{sect:appendix:model_config_and_architecture}

This section details the explicit forward pass formulations, dimensionality, and hyperparameters of the models evaluated in the main text.

\subsection{Input Representation and Standardization}
\label{app:input_representation_and_standardization}
Let the boundary input be represented by $n_{\text{pol}}$ poloidal modes, $n_{\text{tor}}$ toroidal modes, and $n_{\text{arr}}$ arrays (where $n_{\text{arr}} = 2$ under stellarator symmetry, and $4$ otherwise). The boundary is deterministically scaled into a flat vector $\mathbf{x}_{\text{bound}}$.   

The requirement metrics vector $\mathbf{r} = [\normdcc, \normdcp, \normcurvature, \fielderror] \in \mathbb{R}^{4}$ is standardized with robust training-set statistics, the per-metric median $\boldsymbol{\mu}_r$ and
the normal-consistent inter-quartile range $\boldsymbol{\sigma}_r = (\mathbf{q}_{75} - \mathbf{q}_{25})/1.349$,
so that a few extreme instances do not set the scale:
$$
\mathbf{\hat{r}} = \frac{\mathbf{r} - \boldsymbol{\mu}_r}{\max(\boldsymbol{\sigma}_r, 10^{-8})},
$$

With 90 boundary coefficients and four requirements, fixed-shape inputs have dimension 94.
Variable-shape context adds four $\nfp$ and five $\ncoils$ indicators, giving 103.
Per-coil inputs add eight slot indicators and one relative slot position, giving 112.
We denote the resulting input vector by $\mathbf{x}$.

\subsection{Baseline Coil Predictor Formulation}

The plain \mlp{} has an input affine projection, three GELU--affine hidden transformations, and a GELU--affine output projection: five affine layers.
With hidden width $d_h$ and $L=4$ hidden stages,
$$
h_0 = W_{\mathrm{in}}\mathbf{x}+b_{\mathrm{in}},\qquad
h_l = W_l\operatorname{GELU}(h_{l-1})+b_l,\quad l=1,\dots,L-1.
$$
Here $W_{\mathrm{in}}\in\mathbb{R}^{d_h\times d_{\mathrm{in}}}$ and $W_l\in\mathbb{R}^{d_h\times d_h}$.

The final flat output $\mathbf{y} \in \mathbb{R}^{d_{\text{out}}}$ is computed as $\mathbf{y} = W_L \text{GELU}(h_{L-1}) + b_L$. The output dimension encapsulates 3 spatial arrays per coil and 1 scalar current per coil:
$$
d_{\text{out}} = 3 \cdot n_{\text{coils}} \cdot N_{\text{max}} + n_{\text{coils}}
$$

\subsection{Bidirectional Exponential Spectral Scaling}

High-frequency Fourier modes in physical geometries typically exhibit exponentially decaying amplitudes.
Left unscaled, these high-order coefficients would be overwhelmed by the dominant low-frequency components, leading to a spectral bias where the neural network fails to resolve fine-grained geometric details.
To counteract this, we apply a bidirectional exponential spectral scaling transformation prior to flattening the Fourier surfaces and coilsets.

For surface modes $(m,n)$ we use $q=\max(|m|,|n|)$; for coil modes $k$ we use $q=|k|$.
For each spatial coefficient $c$, encoding and decoding are
$$
d=\max(1.2^{-q},10^{-7}),\qquad
\hat c=\frac{c}{R_0d},\qquad c=R_0d\hat c,
$$
where $R_0$ is the boundary's zero-mode radial coefficient.
The transformations condition the network representation.
Training minimizes a normalized squared error between predicted and target coil coefficients and currents after decoding.
Physical feasibility is evaluated separately using the benchmark metrics.

\subsection{Residual Predictor and Split Current Head}
The \resmlp{} begins with $h_0=W_{\mathrm{in}}\mathbf{x}+b_{\mathrm{in}}$, followed by three pre-normalized residual blocks and separate geometry and current heads.
The residual update for $l \in \{1, \dots, L-1\}$ is defined as:
$$
h_l = h_{l-1} + \text{Dropout}\left( W_l \text{GELU}\big(\text{LayerNorm}(h_{l-1})\big) + b_l \right)
$$

The geometric coil parameters $\mathbf{y}_{\text{coils}} \in \mathbb{R}^{d_{\text{coils}}}$ (where $d_{\text{coils}} = 3 \cdot n_{\text{coils}} \cdot N_{\text{max}}$) are predicted via a primary head:
$$
\mathbf{y}_{\text{coils}} = W_{\text{head}} \text{GELU}(h_{L-1}) + b_{\text{head}}
$$

The coil currents $\mathbf{y}_{\text{currents}} \in \mathbb{R}^{n_{\text{coils}}}$ are predicted by a specialized sub-network with hidden dimension $d_c$, conditioned on the predicted geometry and standardized metrics:
$$
\mathbf{c}_{\text{in}} = [\mathbf{y}_{\text{coils}}, \mathbf{\hat{r}}]
$$
$$
h_c = W_{c1} \text{LayerNorm}(\mathbf{c}_{\text{in}}) + b_{c1}
$$
$$
\mathbf{y}_{\text{currents}} = W_{c2} \text{GELU}(h_c) + b_{c2}
$$
Zero initialization gives $I=R_0\mu_I$, where $\mu_I$ is the training-set median of $I/R_0$. The final current-head projection parameters are initialized to zero: $W_{c2} = \mathbf{0}, b_{c2} = \mathbf{0}$.

\subsection{Inference-Time Fourier Truncation}
If a specific Fourier order $k$ is requested at inference, such that $k \le (N_{\text{max}} - 1) / 2$, the full-band spatial arrays ($\mathbf{x}_n, \mathbf{y}_n, \mathbf{z}_n$) are centrally sliced along their final spectral axis.
For a required width $w = 2k + 1$, the padding offset is computed as $p = (N_{\text{max}} - w) / 2$, and the arrays are truncated to the indices $[p : p + w]$. The scalar currents remain unaffected.

\subsection{Hyperparameter Specifications}

The exact configuration and training parameters used across the architectures are detailed in \autoref{tab:hyperparameters:configuration} and \autoref{tab:hyperparameters:training}.

\begin{table}[htbp]
\centering
\caption{Hyperparameter Configuration}
\label{tab:hyperparameters:configuration}
\begin{tabular}{@{}lll@{}}
\toprule
\textbf{Parameter} & \textbf{Value} & \textbf{Description} \\
\midrule
$d_h$ & 512 & Width of the primary MLP trunk.
\\
$L$ & 4 & Four hidden stages; separate output projection. \\
$d_c$ & 64 & Hidden dimension of the specialized currents head.
\\
Dropout & 0.2 & Dropout rate (exclusive to Residual models). \\
$M$ & 5 & Number of members in ensemble variants.
\\
$\alpha$ & 1.2 & Spectral-scaling exponent for bidirectional flat mapping.
\\
$s_{\min}$ & $10^{-7}$ & Minimum floor for spectral scaling.
\\
$p$ & $\infty$ & Norm order used to combine surface mode indices.
\\
$\mathbf{\sigma}_r$ clamp & $10^{-8}$ & Minimum robust scale for requirement standardization.
\\
$N_{\text{coils}}^{\max}$ & 8 & Maximum unique coils per half field period (variable-shape readout). \\
$\mathcal{S}_{\nfp}$ & $\{2,3,4,5\}$ & Supported field periods (variable-shape conditioning). \\
$\mathcal{S}_{\ncoils}$ & $\{4,5,6,7,8\}$ & Supported coil counts (variable-shape conditioning). \\
\bottomrule
\end{tabular}
\end{table}

\begin{table}[htbp]
\centering
\caption{Training Hyperparameters}
\label{tab:hyperparameters:training}
\begin{tabular}{@{}lp{6cm}l@{}}
\toprule
\textbf{Parameter} & \textbf{Value} \\
\midrule
Seeds & 0, 1, 2 \\
Training budget & Up to 600,000 updates with 1,000-step warmup. \\
Batch / selection & Train batch 16, evaluation batch 256, and evaluation every 500 updates. \\
Stopping Patience & 100,000 updates with best checkpoint retained. \\
Optimizer & AdamW with learning rate schedule \num{1e-3} to \num{1e-5}, and weight decay \num{1e-4}, and global norm clip 1.0. \\
\bottomrule
\end{tabular}
\end{table}
\subsection{Variable-Shape Topology Conditioning}

\label{sect:appendix:variable_shape}
The variable-shape models (\maskedmlp{}, \percoilmlp{}, and their ensembles)
share the trunk, spectral scaling, and currents handling of the fixed-shape
models, and additionally condition on the discrete topology. The number of field
periods and the coil count are each supplied as a one-hot vector over the
supported values $\nfp \in \mathcal{S}_{\nfp} = \{2,3,4,5\}$ and
$\ncoils \in \mathcal{S}_{\ncoils} = \{4,5,6,7,8\}$, concatenated to the
standardized input. Both architectures emit up to
$N_{\text{coils}}^{\max} = 8$ unique coils per half field period; at inference
the output is trimmed to the requested $\ncoils$ (the extra slots are discarded).

\paragraph{Masked predictor.}
The \maskedmlp{} reuses the basic \mlp{} trunk but sizes its readout for the
maximum coil count,
$$
d_{\text{out}} = 3 \cdot N_{\text{coils}}^{\max} \cdot N_{\max} + N_{\text{coils}}^{\max}.
$$
All $N_{\text{coils}}^{\max}$ slots are produced in a single forward pass, and
slots beyond the requested $\ncoils$ are masked out by the caller.

\paragraph{Per-coil predictor.}
The \percoilmlp{} applies a single weight-shared MLP to each coil slot in turn.
Slot $j \in \{0, \dots, N_{\text{coils}}^{\max}-1\}$ receives the shared design
context (flattened boundary, standardized requirements, and the $\nfp$/$\ncoils$
one-hots) concatenated with a slot descriptor: the one-hot slot index and the
relative coil-slot position $(j + \tfrac{1}{2})/\ncoils$. Each slot emits its three
spatial Fourier arrays and a scalar current, so a single parameter set
represents any coil count.

\paragraph{Ensembles.}
Each ensemble contains five independently initialized networks trained on shared minibatches.
We average their individual supervised losses during training and their predicted coil coefficients and currents at inference.
We repeat each single-model or ensemble run with three seeds and report variation across those runs. Because the flat-to-coilset decoding is affine,
this equals averaging the members' flat outputs.

%%%%%%%%%%%%%%%%%%%%%%%%%%%%%%%%%%%%%%%%%%%%%%%%%%%%%%%%%%%%

% \newpage
% \input{checklist.tex}

\end{document}